\documentclass{article}

\usepackage{microtype}
\usepackage{graphicx}
\usepackage{subfigure}
\usepackage{booktabs} 
\usepackage{multirow}
\usepackage{manyfoot}

\usepackage{hyperref}

\usepackage[accepted]{icml2024}

\usepackage{amsmath}
\usepackage{amssymb}
\usepackage{mathtools}
\usepackage{amsthm}
\usepackage{graphicx}

\usepackage{tabularx}

\usepackage[capitalize,noabbrev]{cleveref}
\newcommand{\oursfull}{DoRA: Weight-Decomposed Low-Rank Adaptation}
\newcommand{\oursfullbold}{Weight-\textbf{D}ecomposed L\textbf{o}w-\textbf{R}ank \textbf{A}daptation}
\newcommand{\ours}{DoRA}
\newcommand{\ourss}{DVoRA}

\theoremstyle{plain}

\theoremstyle{definition}

\theoremstyle{remark}

\usepackage[textsize=tiny]{todonotes}

\icmltitlerunning{DoRA: Weight-Decomposed Low-Rank Adaptation}

\begin{document}

\twocolumn[
\icmltitle{\oursfull}

\begin{icmlauthorlist}
\icmlauthor{Shih-Yang Liu}{yyy,comp}
\icmlauthor{Chien-Yi Wang}{yyy}
\icmlauthor{Hongxu Yin}{yyy}
\icmlauthor{Pavlo Molchanov}{yyy}
\icmlauthor{Yu-Chiang Frank Wang}{yyy}
\icmlauthor{Kwang-Ting Cheng}{comp}
\icmlauthor{Min-Hung Chen}{yyy}
\end{icmlauthorlist}

\icmlaffiliation{yyy}{NVIDIA}
\icmlaffiliation{comp}{HKUST}

\icmlcorrespondingauthor{Shih-Yang Liu}{shihyangl@nvidia.com, sliuau@connect.ust.hk}
\icmlcorrespondingauthor{Min-Hung Chen}{minhungc@nvidia.com}
\icmlkeywords{Machine Learning, ICML}

\vskip 0.3in
]



\printAffiliationsAndNotice{} 

\begin{abstract}
Among the widely used parameter-efficient fine-tuning (PEFT) methods, LoRA and its variants have gained considerable popularity because of avoiding additional inference costs. However, there still often exists an accuracy gap between these methods and full fine-tuning (FT). In this work, we first introduce a novel weight decomposition analysis to investigate the inherent differences between FT and LoRA. Aiming to resemble the learning capacity of FT from the findings,
we propose Weight-\textbf{D}ecomposed L\textbf{o}w-\textbf{R}ank \textbf{A}daptation (\textbf{DoRA}). DoRA decomposes the pre-trained weight into two components, \textit{magnitude} and \textit{direction}, for fine-tuning, specifically employing LoRA for directional updates to efficiently minimize the number of trainable parameters. By employing DoRA, we enhance both the learning capacity and training stability of LoRA while avoiding any additional inference overhead. DoRA consistently outperforms LoRA on fine-tuning LLaMA, LLaVA, and VL-BART on various downstream tasks, such as commonsense reasoning, visual instruction tuning, and image/video-text understanding. Code is available at \url{https://github.com/NVlabs/DoRA}.

\end{abstract}
\section{Introduction}
\begin{figure}[th]
\vskip 0.2in
\begin{center}
\centerline{\includegraphics[width=\columnwidth]{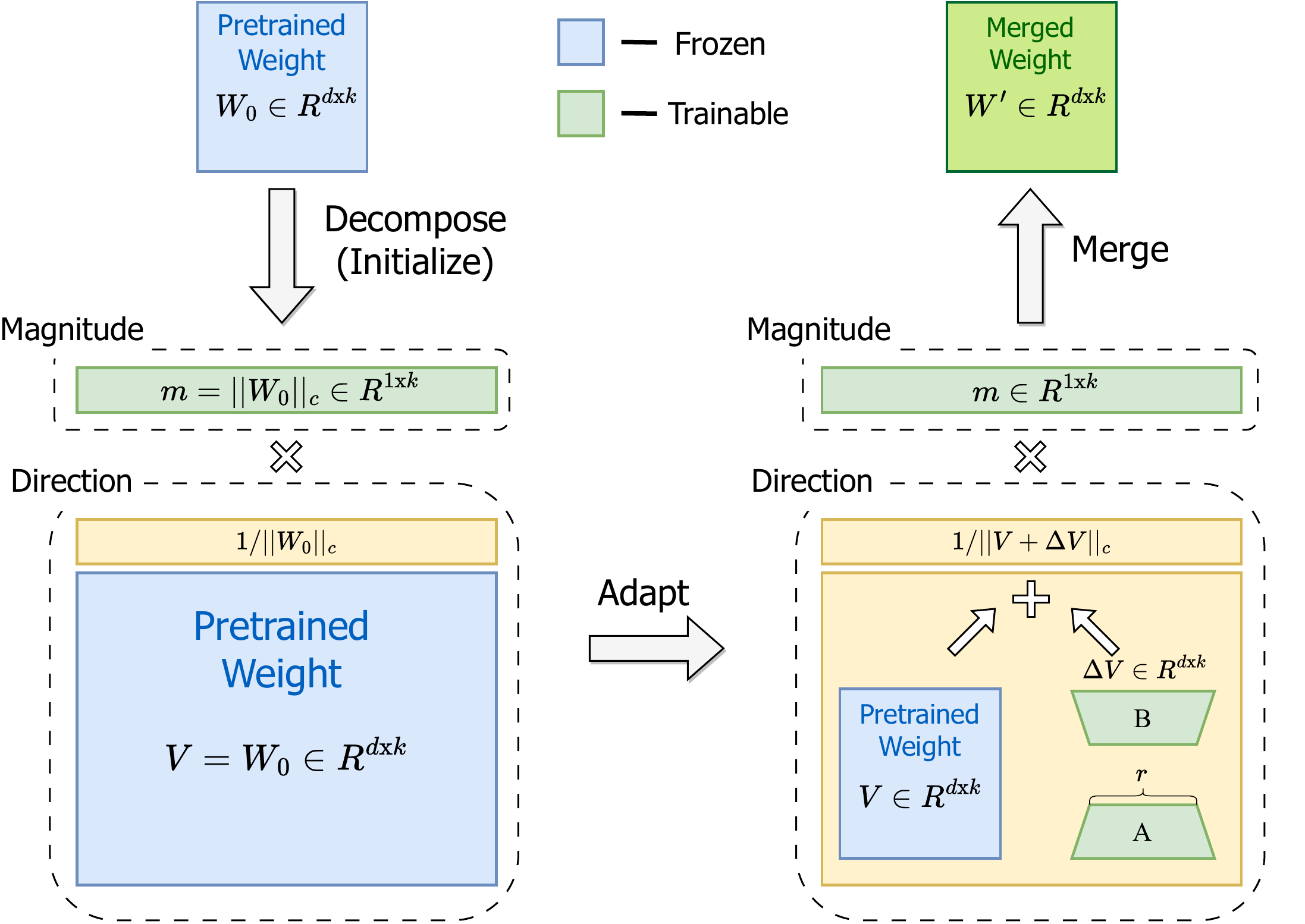}}
\caption{An overview of our proposed \ours, which decomposes the pre-trained weight into \textit{magnitude} and \textit{direction} components for fine-tuning, especially with LoRA to efficiently update the direction component. Note that $||\cdot||_{c}$ denotes the vector-wise norm of a matrix across each column vector.
}
\label{fig:MADoRA}
\end{center}
\vskip -0.2in
\end{figure}
Models that are pre-trained with extensive general domain datasets have demonstrated remarkable generalization abilities, significantly benefiting a wide array of applications, from natural language processing (NLP) tasks \cite{qin2023chatgpt, alpaca} to multi-modal tasks \cite{li2022blip, liu2023llava}. To tailor these general models for specific downstream tasks, \textit{full fine-tuning (FT)} is commonly employed, involving the retraining of all model parameters. Nevertheless, as the size of models and datasets expand in scale, the expense associated with fine-tuning the entire model becomes prohibitively large.

To address this issue, parameter-efficient fine-tuning (PEFT) methods \cite{houlsby2019series} have been introduced to fine-tune the pre-trained models with only a minimal number of parameters. Among these, LoRA \cite{hu2022lora}, which does not change the model architecture, has become notably popular for its simplicity and efficacy. Nevertheless, there is still a capacity gap between LoRA and FT, which is often attributed to the limited number of trainable parameters without further exploration of other underlying causes~\cite{hu2022lora, kopiczko2023vera}.

Drawing on Weight Normalization~\cite{salimans2016weight}, which achieves faster convergence via improving the conditioning of the gradient with weight reparameterization, we introduce a novel weight decomposition analysis that initially reparameterizes model weights into magnitude and directional components, subsequently examining the changes in magnitude and direction introduced by LoRA and FT. Our analysis reveals that LoRA and FT exhibit markedly distinct patterns of updates, leading us to surmise that these variations mirror the learning capability of each method. Inspired by our findings, we propose \oursfullbold~(\textbf{\ours}), which begins by decomposing the pre-trained weight into its magnitude and directional components, then fine-tunes both. Given the substantial size of the directional component in terms of parameters, we exploit LoRA for the directional adaptation to enable efficient fine-tuning, as illustrated in Figure.\ref{fig:MADoRA}.
Moreover, by showing a learning behavior similar to FT both empirically and mathematically, suggesting a learning capacity closely resembling FT, we have validated \ours~across a wide variety of tasks, from NLP to Vision-Language, and over various backbones, including LLM and LVLM. The experimental results show that \ours~consistently outperforms LoRA without sacrificing inference efficiency, such as commonsense reasoning (\textbf{+3.7}/\textbf{+1.0} on LLaMA-7B/13B, \textbf{+2.9} on LLaMA2-7B, and \textbf{+4.4} on LLaMA3-8B), visual instruction tuning (\textbf{+0.6} on LLaVA-7B), and image/video-text understanding (\textbf{+0.9}/\textbf{+1.9} on VL-BART). 


The summary of our contributions is as follows:
\begin{itemize}
\item We introduce \ours, a novel PEFT method that incorporates weight decomposition, achieving a learning capacity closely resembling FT without any additional inference latency over LoRA.
\item We introduce a novel weight decomposition analysis to uncover the fundamental differences in the learning patterns of FT and different PEFT methods.
\item \ours~consistently surpasses LoRA on various tasks, from NLP to Vision-Language benchmarks and across various backbones, including LLM and LVLM.
\end{itemize}


\section{Related Works}
\textbf{Parameter-Efficient Fine-Tuning (PEFT)} methods are designed to reduce the high expense of fine-tuning large-scale models. They achieve this by training a relatively small subset of parameters, compared to the total number of parameters, for adapting to downstream tasks. Existing PEFT methods can be divided into three categories. The first category is referred to as \textit{Adapter-based} methods, which involve introducing additional trainable modules into the original frozen backbone, such as \cite{houlsby2019series, he2022parallel, mahabadi2021parameterefficient, mahabadi2021compacter}. For example, \cite{houlsby2019series} proposes adding linear modules in sequence to the existing layer, whereas \cite{he2022parallel} advocates for integrating these modules in parallel with the original layer to enhance performance. The second category is \textit{Prompt-based} methods. These methods add extra soft tokens (prompts) to the initial input and focus solely on fine-tuning these trainable vectors, as seen in works like \cite{lester2021power, razdaibiedina2023residual, wang2023nonintrusive}. However, these approaches typically face challenges due to their sensitivity to initialization, affecting their overall effectiveness. These first two categories, whether altering the model's input or architecture, result in increased inference latency compared to the baseline model.

\textbf{LoRA~\cite{hu2022lora} and its variants} are among the third category of PEFT, notable for not adding any extra inference burden. These methods apply low-rank matrices to approximate weight changes during fine-tuning and can merge with pre-trained weights prior to inference. For example, \cite{zhang2023adalora} employs SVD decomposition and prunes less significant singular values for more efficient updates. \cite{hyeonwoo2023fedpara} focuses on low-rank Hadamard product for federated learning. \cite{qiu2023controlling, liu2023butterfly} exploit orthogonal factorization in fine-tuning diffusion models. \cite{renduchintala2023tied} uses weight tying to further reduce the trainable parameters. \cite{yeh2023navigating} introduces a unified LoRA family framework for Stable diffusion. \cite{ponti2022combining} chooses different combinations of LoRAs from the inventory with a routing function for different tasks. \cite{kopiczko2023vera} implements learnable scaling vectors to adjust a shared pair of frozen random matrices across layers. Our research also falls within this third category, and we validate the efficacy of our proposed method alongside LoRA and its variants through comprehensive experimentation.

\section{Pattern Analysis of LoRA and FT}
\subsection{Low-Rank Adaptation (LoRA)}
Building upon the hypothesis that updates made during the fine-tuning exhibit a low ``intrinsic rank", LoRA ~\cite{hu2022lora} proposes using the product of two low-rank matrices to update the pre-trained weights incrementally. For a pre-trained weight matrix $W_0 \in \mathbb{R}^{d \times k}$, LoRA models the weight update $\Delta W \in \mathbb{R}^{d \times k}$ utilizing a low-rank decomposition, expressed as $BA$, where $B\in \mathbb{R}^{d \times r}$ and $A\in \mathbb{R}^{r \times k}$ represent two low-rank matrices, with $r \ll min(d,k)$. Consequently, the fine-tuned weight $W'$ can be represented as:
\begin{equation}
W' = W_0 + \Delta W = W_0 + \underline{BA}
\label{eq:lora}
\end{equation}
where $W_0$ remains static during the fine-tuning process, and the underlined parameters are being trained. The matrix $A$ is initialized with uniform Kaiming distribution~\cite{he2015delving}, while $B$ is initially set to zero, resulting in $\Delta W = BA$ being zero at the start of training. Notably, this decomposition of $\Delta W$ can be substituted with other LoRA variants, such as VeRA~\cite{kopiczko2023vera}.
Additionally, based on Eq.~(\ref{eq:lora}), we can merge the learned $\Delta W$ with the pre-trained weight $W_0$ and obtain $W'$ in advance of deployment, and given that both $W'$ and $W_0$ both fall within the dimensionality of $\mathbb{R}^{d \times k}$, LoRA and its related variants do not introduce any extra latency during the inference compared to the original model.

\begin{figure*}[ht]
\vskip 0.2in
\begin{center}
\centerline{\includegraphics[width=2.1\columnwidth]{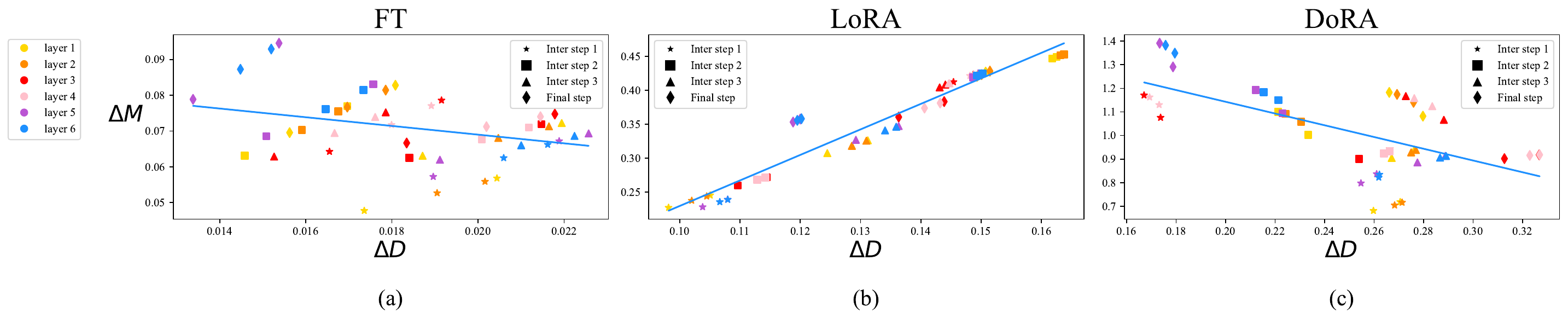}}
\caption{Magnitude and direction updates of (a) FT, (b) LoRA, and (c) \ours~of the query matrices across different layers and intermediate steps. Different markers represent matrices of different training steps and different colors represent the matrices of each layer.}
\label{fig:q_proj_analysis}
\end{center}
\vskip -0.2in
\end{figure*}

\subsection{Weight Decomposition Analysis}

The study presented in LoRA~\cite{hu2022lora} suggests that LoRA can be considered a general approximation of full fine-tuning. By gradually increasing the rank $r$ of LoRA to align with the rank of pre-trained weights, LoRA can attain a level of expressiveness akin to that of FT. Consequently, many previous studies have attributed the discrepancy in accuracy between LoRA and FT primarily to the limited number of trainable parameters, often without further analysis~\cite{hu2022lora, kopiczko2023vera}. Drawing inspiration from Weight Normalization~\cite{salimans2016weight}, which reparameterizes the weight matrix into magnitude and direction for accelerating optimization, we introduce an innovative weight decomposition analysis. Our analysis restructures the weight matrix into two separate components, \textit{magnitude} and \textit{direction}, to reveal the inherent differences in LoRA and FT learning patterns.

\textbf{Analysis Method: }This analysis examines the updates in both magnitude and direction of the LoRA and FT weights relative to the pre-trained weights to reveal the fundamental differences in the learning behaviors of both. The weight decomposition of $W \in \mathbb{R}^{d \times k}$ can be formulated as:
\begin{equation}
W = m\frac{V}{||V||_{c}} = ||W||_{c}\frac{W}{||W||_{c}}
\label{eq:weight_decomposition}
\end{equation}
where $m \in \mathbb{R}^{1 \times k}$ is the magnitude vector, $V \in \mathbb{R}^{d \times k}$ is the directional matrix, with $||\cdot||_{c}$ being the vector-wise norm of a matrix across each column. This decomposition ensures that each column of $V/||V||_{c}$ remains a unit vector, and the corresponding scalar in $m$ defines the magnitude of each vector.


For our weight decomposition analysis, we select the VL-BART model fine-tuned on four image-text tasks as outlined in \cite{sung2022vladapter} for a case study. Following \cite{sung2022vladapter}, which applies LoRA only to the query/value weight matrix in the self-attention module. We decompose the pre-trained weight $W_0$, the full fine-tuned weight $W_{\text{FT}}$, and the merged LoRA weight $W_{\text{LoRA}}$ of query/value weight matrix using Eq.~(\ref{eq:weight_decomposition}). The magnitude and directional variations between $W_0$ and $W_{\text{FT}}$ can be defined as follows:
\begin{align}
\label{eq:weight_decomposition_difference}
\Delta M_{\text{FT}}^{t} = \frac{\sum_{n = 1}^{k} |m_{\text{FT}}^{n,t} - m_0^{n}|}{k}\\
\label{eq:weight_decomposition_difference2}
\Delta D_{\text{FT}}^{t} = \frac{\sum_{n = 1}^{k} (1- \mathbf{cos}(V_{\text{FT}}^{n,t}, W_{0}^{n}))}{k}
\end{align}


Here, $\Delta M_{\text{FT}}^{t}$ and and $\Delta D_{\text{FT}}^{t}$ represent the magnitude difference and directional difference between $W_0$ and $W_{\text{FT}}$ at $t$ training step respectively, with $\mathbf{cos}(\cdot,\cdot)$ being the cosine similarity function. $M_{\text{FT}}^{n,t}$ and $M_0^{n}$ are the $n^{th}$ scalars in their respective magnitude vectors, while $V_{\text{FT}}^{n,t}$ and $W_{0}^{n}$ are the $n^{th}$ columns in $V_{\text{FT}}^{t}$ and $W_{0}$. The magnitude and directional differences between $W_{\text{LoRA}}$ and $W_0$ are calculated similarly, as per Eq.~(\ref{eq:weight_decomposition_difference}) and Eq.~(\ref{eq:weight_decomposition_difference2}). We select checkpoints from four different training steps for analysis, comprising three intermediate steps and the final checkpoint from both FT and LoRA, and we perform weight decomposition analysis on each of these checkpoints to determine the $\Delta M$ and $\Delta D$ throughout different layers.

\textbf{Analysis Results:} Figure \ref{fig:q_proj_analysis} (a) and (b) illustrate the alterations in the query weight matrix of FT and LoRA, with each point representing a ($\Delta D^{t}$, $\Delta M^{t}$) pair from query weight matrices across different layers and training steps. Similarly, Figure \ref{fig:v_proj_analysis} in the appendix displays the value weight matrix modifications. It is noticeable that LoRA exhibits a consistent positive slope trend across all the intermediate steps, signifying a proportional relationship between the changes in direction and magnitude. In contrast, the FT displays a more varied learning pattern with a relatively negative slope. This distinction between FT and LoRA likely mirrors their respective learning capability. While LoRA tends to either increase or decrease the magnitude and direction updates proportionally, it lacks the nuanced capability for more subtle adjustments. Specifically, LoRA does not show proficiency in executing slight directional changes alongside more significant magnitude alterations, or vice versa, a feature more characteristic of the FT method. We suspect that such limitation of LoRA might stem from the challenge of concurrent learning both magnitude and directional adaptation, which could be overly complex for LoRA. Consequently, in this work, we aim to propose a variant of LoRA that exhibits a learning pattern more closely resembling that of FT, and can improve the learning capacity over LoRA. 


\section{Method} 
\subsection{\oursfullbold}
\label{sec:MADoRA_method}
Drawing from the insights of our weight decomposition analysis, we introduce \oursfullbold~(\textbf{\ours}). \ours~initially decomposes the pre-trained weight into its magnitude and directional components and finetunes both of them. Because the directional component is large in terms of parameter numbers, we further decompose it with LoRA for efficient finetuning. 

Our intuitions are two-fold. Firstly, we believe that limiting LoRA to concentrate exclusively on directional adaptation while also allowing the magnitude component to be tunable simplifies the task compared to the original approach, where LoRA is required to learn adjustments in both magnitude and direction. Secondly, the process of optimizing directional updates is made more stable through weight decomposition, which we delve into more thoroughly in Section.\ref{sec:gradient}. It is important to highlight that the main distinction between \ours~and weight normalization \cite{salimans2016weight} lies in their training approaches. Weight normalization trains both components from scratch, making the method sensitive to different initializations. Conversely, \ours~avoids such initialization concerns since both components begin with pre-trained weights. We initialize \ours~with pre-trained weight $W_0$ as outlined in Eq.~(\ref{eq:weight_decomposition}), where $m = ||W_0||_{c}$ and $V = W_0$ after initialization. We then keep $V$ frozen and $m$ a trainable vector. The directional component is then updated through LoRA. 
\ours~can be formulated similar to Eq.~(\ref{eq:lora}) as: 
\begin{equation}
W'= \underline{m}\frac{V+\Delta V}{||V+\Delta V||_{c}} = \underline{m}\frac{W_0+\underline{BA}}{||W_0+\underline{BA}||_{c}}
\label{eq:MADoRA}
\end{equation}
where $\Delta V$ is the incremental directional update learned by multiplying two low-rank matrices $B$ and $A$, and the underlined parameters denote the trainable parameters. The matrices $B\in \mathbb{R}^{d \times r}$ and $A\in \mathbb{R}^{r \times k}$ are initialized in line with LoRA's strategy to ensure that $W'$ equals $W_0$ before the finetuning. Furthermore, \ours~can be merged with the pre-trained weight before inference, thereby not introducing any additional latency. 

We visualize the magnitude and directional differences of the query weight matrix between the merged \ours~weight and $W_0$ in the same setting as for FT and LoRA in Figure~\ref{fig:q_proj_analysis} (c) and leave the visualization of the value weight matrix in the appendix. From the regression line for $(\Delta D, \Delta M)$ of both \ours~and FT, we reveal that in contrast to LoRA's pattern, \ours, and FT are characterized by a distinct negative slope. We reason that FT tends towards a negative slope because pre-trained weights already possess substantial knowledge suitable for various downstream tasks. Therefore, when provided with adequate learning capacity, having a larger magnitude or direction alteration alone is sufficient enough for downstream adaptation. We additionally compute the correlation between $\Delta D$ and $\Delta M$ for FT, LoRA, and \ours, and we found that both FT and \ours~exhibit negative correlation values of -0.62 and -0.31, respectively. In contrast, LoRA shows a positive correlation with a value of 0.83. In conclusion, the fact that \ours~demonstrates the ability to make only substantial directional adjustments with relatively minimal changes in magnitude or the reverse while showing learning patterns closer to FT's signifies its superior learning capacity over LoRA.



\subsection{Gradient Analysis of \ours}
\label{sec:gradient}
In this section, we first derive the gradient of \ours~and illustrate how our proposed decomposition benefits the optimization of $\Delta V$. Subsequently, we analyze from the gradient's perspective to explicate the learning pattern of \ours, which tends to have a negative slope. 

From Eq.~(\ref{eq:MADoRA}), we can obtain the gradient of Loss $\mathcal{L}$ with respect to $m$ and $V' = V+\Delta V$ as:
\begin{align}
\label{eq:MADoRA_v_gradient}
\nabla_{V'} \mathcal{L} &= \frac{m}{||V'||_{c}}\left( I - \frac{V'V'^{\mathbf{T}}}{||V'||_{c}^2}  \right) \nabla_{W'} \mathcal{L} \\
\label{eq:MADoRA_m_gradient}
\nabla_m \mathcal{L} &=\frac{\nabla_{W'} \mathcal{L} \cdot V'}{||V'||_{c}}
\end{align}
Eq.~(\ref{eq:MADoRA_v_gradient}) reveals that the weight gradient $\nabla_{W'} \mathcal{L}$ is scaled by $m/||V'||_{c}$ and is projected away from the current weight matrix. These two effects contribute to aligning the gradient's covariance matrix more closely with the identity matrix, which is advantageous for optimization \cite{salimans2016weight}. Additionally, given that $V' = V + \Delta V$, the gradient $\nabla_{V'} L$ is equivalent to $\nabla_{\Delta V} L$. Therefore, the optimization benefits derived from this decomposition are fully transferred to $\Delta V$, enhancing the learning stability of LoRA.

We can gain further insight into the learning pattern of \ours~by referring to Eq.~(\ref{eq:MADoRA_m_gradient}). In the subsequent discussion, we represent vectors using lower-case letters instead of the previous matrix form notation. Consider $w'' = w' + \Delta w$ as the parameter update for a weight vector, where $\Delta w \propto \nabla_{w'} \mathcal{L}$. In two hypothetical update scenarios, $S1$ and $S2$, $S1$ involves a smaller directional update ($\Delta D_{S1}$), while $S2$ involves a larger one ($\Delta D_{S2}$). Assuming $||\Delta w_{S1}|| = ||\Delta w_{S2}||$, and at time 0, we have $\Delta v = 0$ and $v' = v$. From $\Delta D_{S1} < \Delta D_{S2}$, it follows that $|\mathbf{cos}(\Delta w_{S1}, w')| > |\mathbf{cos}(\Delta w_{S2}, w')|$. Since $\Delta w \propto \nabla_{w'} \mathcal{L}$, it implies $|\mathbf{cos}(\nabla_{w'}^{S1} \mathcal{L}, w')| > |\mathbf{cos}(\nabla_{w'}^{S2} \mathcal{L}, w')|$. From Sec~\ref{sec:MADoRA_method}, with $v$ initialized as $v_0$ and $w' = w_0$ at time 0, we get $|\mathbf{cos}(\nabla_{w'} \mathcal{L}, w')| = |\mathbf{cos}(\nabla_{w'} \mathcal{L}, v')| = |\mathbf{cos}(\nabla_{w'} \mathcal{L}, v)|$. Using the cosine similarity equation with $\Delta v = 0$:
\begin{equation}
cos(\nabla_{w'} \mathcal{L}, v') = cos(\nabla_{w'} \mathcal{L}, v) =  \frac{\nabla_{w'} \mathcal{L} \cdot v}{||\nabla_{w'} \mathcal{L}||||v||}
\end{equation}
denote $m_{*}$ as the magnitude scalar of vector $w'$
then Eq.~(\ref{eq:MADoRA_m_gradient}) w.r.t $m_{*}$ can be rewritten to:
\begin{equation}
\nabla_{m_{*}} \mathcal{L} = \frac{\nabla_{w'} \mathcal{L} \cdot v'}{||v'||} = ||\nabla_{w'} \mathcal{L}|| \cdot cos(\nabla_{w'} \mathcal{L}, v)
\end{equation}

Given that $||\Delta w_{S1}|| = ||\Delta w_{S2}||$ for $S1$ and $S2$, and $||\nabla_{w'}^{S1} \mathcal{L}|| = ||\nabla_{w'}^{S2} \mathcal{L}||$. Therefore, with:
\begin{equation}
    ||\nabla_{w'}^{S1} \mathcal{L}|| \cdot |cos(\nabla_{w'}^{S1} \mathcal{L}, v)| > ||\nabla_{w'}^{S2} \mathcal{L}|| \cdot |cos(\nabla_{w'}^{S2} \mathcal{L}, v)|
\end{equation}
it can be inferred that $|\nabla_{m_{*}}^{S1} \mathcal{L}| > |\nabla_{m_{*}}^{S2} \mathcal{L}|$ which indicate that $S1$ has larger magnitude updates over $S2$ while having smaller directional alteration than that of $S2$. Our conclusion generally holds in practice, as evidenced by Figure~\ref{fig:q_proj_analysis} (c). Consequently, we have effectively shown how \ours~can be utilized to adjust the learning pattern, diverging from that of LoRA and aligning more closely with the pattern of FT.

\subsection{Reduction of Training Overhead}
In Eq.~(\ref{eq:lora}), the gradients of $W'$ and $\Delta W$ are the same. However, with \ours, which redirects the low-rank adaptation towards the directional component, the gradient of the low-rank updates differs from that of $W'$, as illustrated in Eq.~(\ref{eq:MADoRA_v_gradient}). This divergence necessitates extra memory during backpropagation. To address this, we suggest treating $||V + \Delta V||_{c}$ in Eq.~(\ref{eq:MADoRA}) as a constant, thereby detaching it from the gradient graph. This means that while $||V + \Delta V||_{c}$ dynamically reflects the updates of $\Delta V$, it won't receive any gradient during backpropagation. With this modification, the gradient w.r.t $m$ remains unchanged, and $\nabla_{V'} \mathcal{L}$ is redefined as:
\begin{equation}
\nabla_{V'} \mathcal{L} = \frac{m}{C} \nabla_{W'} \mathcal{L} \text{ where } C = ||V'||_{c}
\end{equation}
This approach reduces the gradient graph memory consumption drastically without a noticeable difference in accuracy. We conduct an ablation study to evaluate the impact of the proposed modification on fine-tuning LLaMA-7B and VL-BART. 
The results indicate that the modification leads to a training memory reduction of approximately 24.4\% in fine-tuning LLaMA and 12.4\% in VL-BART. Furthermore, the accuracy of \ours~with the modification remains unchanged for VL-BART and shows a negligible difference of only 0.2 compared to \ours~without the modification on LLaMA.
For a comprehensive comparison of training memory usage and accuracy differences, please see Table \ref{tab:training_cost} in the appendix. Consequently, all subsequent experiments with \ours~incorporate this adjustment. 

\section{Experiments}
We conduct a variety of experiments to showcase the efficacy of \ours~on various tasks including language, image, and video domains. Firstly, we evaluate \ours~against several Parameter-Efficient Fine-Tuning (PEFT) methods by fine-tuning LLaMA-7B/13B, LLaMA2-7B, and LLaMA3-8B on commonsense reasoning tasks. Subsequently, we extend from single modality to multimodality. We compare \ours~with LoRA across multi-task image/video-text understanding tasks using VL-BART and visual instruction tuning with LLaVA-1.5-7B. Following this, we explore the compatibility of \ours~with LoRA and VeRA~\cite{kopiczko2023vera} for instruction-tuning on LLaMA-7B and LLaMA2-7B. Furthermore, we perform a series of ablation studies to illustrate that \ours~surpasses LoRA in performance, irrespective of the number of fine-tuning training samples and rank variations.  Lastly, We analyze the tuning granularity of \ours, and show that \ours~can achieve better accuracy than LoRA with fewer trainable parameters by selectively updating only the directional components of certain modules. 

\subsection{Commonsense Reasoning}
\label{sec:llama_commonsense}
\begin{table*}[t]
\caption{Accuracy comparison of LLaMA 7B/13B, LLaMA2 7B, and LLaMA3 8B with various PEFT methods on eight commonsense reasoning datasets. Results of all the baseline methods on LLaMA 7B/13B are taken from \cite{hu2023llmadapters}. Results of LoRA on LLaMA2 7B and LLaMA3 8B are obtained using the hyperparameters  described in \cite{hu2023llmadapters}. $\text{\ours}^{\dagger}$: the adjusted version of \ours~with the rank halved.
}
\vskip 0.1in
\setlength{\tabcolsep}{1.2mm}
\centering
\resizebox{0.99\textwidth}{!}{
\begin{tabular}{ccccccccccccc}
\toprule
\textbf{Model} & \textbf{PEFT Method} & \# \textbf{Params (\%)} & \textbf{BoolQ} & \textbf{PIQA}&\textbf{SIQA}& \textbf{HellaSwag} & \textbf{WinoGrande}& \textbf{ARC-e} & \textbf{ARC-c} & \textbf{OBQA} & \textbf{Avg.} \\ \hline
ChatGPT & - & - & 73.1 & 85.4 & 68.5 & 78.5 & 66.1 & 89.8 & 79.9 & 74.8 & 77.0 \\ \hline
\multirow{6}{*}{LLaMA-7B}    & Prefix & 0.11 & 64.3 & 76.8 & 73.9 & 42.1 & 72.1 & 72.9 & 54.0 & 60.6 & 64.6 \\ 
    & Series & 0.99 & 63.0 & 79.2 & 76.3 & 67.9 & 75.7 & 74.5 & 57.1 & 72.4 & 70.8 \\ 
 & Parallel & 3.54 & 67.9 & 76.4 & 78.8 & 69.8 & 78.9 & 73.7 & 57.3 & 75.2 & 72.2 \\ 
                                & LoRA & 0.83 & 68.9 & 80.7 & 77.4 & 78.1 & 78.8 & 77.8 & 61.3 & 74.8 & 74.7 \\ 
                                 & $\text{\ours}^{\dagger}$~(Ours)  & 0.43 & 70.0 & 82.6 & 79.7 & 83.2 & 80.6 & 80.6 & 65.4 & 77.6 & \textbf{77.5} \\ 
                                 & \ours~(Ours)  & 0.84 & 69.7 &	83.4 &	78.6 &	87.2 &	81.0 &	81.9 &	66.2 &	79.2 &	\textbf{78.4} \\ \hline
\multirow{6}{*}{LLaMA-13B}    & Prefix & 0.03 & 65.3 & 75.4 & 72.1 & 55.2 & 68.6 & 79.5 & 62.9 & 68.0 & 68.4 \\ 
    & Series & 0.80 & 71.8 & 83 & 79.2 & 88.1 & 82.4 & 82.5 & 67.3 & 81.8 & 79.5 \\ 
 & Parallel & 2.89 & 72.5 & 84.9 & 79.8 & 92.1 & 84.7 & 84.2 & 71.2 & 82.4 & 81.4 \\ 
                                & LoRA & 0.67 & 72.1 & 83.5 & 80.5 & 90.5 & 83.7 & 82.8 & 68.3 & 82.4 & 80.5 \\ 
                                 & $\text{\ours}^{\dagger}$~(Ours) & 0.35 & 72.5 & 85.3 & 79.9 & 90.1 & 82.9 & 82.7 & 69.7 & 83.6 & \textbf{80.8}\\ 
                                 & \ours~(Ours)& 0.68 & 72.4 & 84.9 & 81.5 & 92.4 & 84.2 & 84.2 & 69.6 & 82.8 & \textbf{81.5} \\ \hline
\multirow{3}{*}{LLaMA2-7B}    & LoRA & 0.83 & 69.8&	79.9&	79.5&	83.6&	82.6&	79.8&	64.7&	81.0&	77.6 \\ 
& $\text{\ours}^{\dagger}$~(Ours) & 0.43 & 72.0	&83.1	&79.9&	89.1&	83.0	&84.5&	71.0	&81.2&	\textbf{80.5}\\ 
    & DoRA~(Ours) & 0.84 & 71.8 &	83.7&	76.0&	89.1&	82.6&	83.7&	68.2&	82.4&	\textbf{79.7} \\ \hline
\multirow{3}{*}{LLaMA3-8B}    & LoRA & 0.70 &70.8&	85.2&	79.9&	91.7&	84.3&	84.2&	71.2&	79.0&	80.8 \\ 
& $\text{\ours}^{\dagger}$~(Ours) & 0.35 & 74.5 &	88.8&	80.3&	95.5&	84.7&	90.1&	79.1&	87.2&	\textbf{85.0}\\ 
    & DoRA~(Ours) & 0.71 & 74.6&	89.3&	79.9&	95.5&	85.6&	90.5&	80.4&	85.8&	\textbf{85.2} \\ \hline
\end{tabular}}
\label{tab:llama_commonsense}
\vskip -0.1in
\end{table*}

We evaluate \ours~against LoRA and several baseline methods which include \textit{Prompt learning (Prefix)}~\cite{li2021prefixtuning}, \textit{Series adapter (Series)}~\cite{houlsby2019series}, and \textit{Parallel adapter (Parallel)}~\cite{he2022parallel}
on LLaMA-7B/13B~\cite{touvron2023llama} for commonsense reasoning tasks. We also include ChatGPT’s accuracy obtained with gpt-3.5-turbo API using a zero-shot Chain of Thought \cite{chatgptdoc, wei2023chainofthought}.

The commonsense reasoning tasks comprise 8 sub-tasks, each with a predefined training and testing set. We follow the setting of \cite{hu2023llmadapters} and amalgamate the training datasets from all 8 tasks to create the final training dataset and conduct evaluations on the individual testing dataset for each task. 
To ensure a fair comparison, we initially fine-tuned models with \ours~following the LoRA configuration, maintaining the same rank while adjusting only the learning rate.
The marginal increase of 0.01\% in the number of trainable parameters for \ours~over LoRA, as detailed in Table~\ref{tab:llama_commonsense}, arises from the inclusion of learnable magnitude components (parameter of size ${1 \times k}$). Then, we further halve the rank used in \ours~compared to LoRA and denote this adjusted configuration as $\text{\ours}^{\dagger}$. See Table \ref{tab:llama_commonsense_hyperparameters} for details on the hyperparameters used. 

Table~\ref{tab:llama_commonsense} demonstrates that \ours~consistently surpasses all baseline methods across both LLaMA-7B/13B, LLaMA2-7B and LLaMA3-8B. Notably, in the LLaMA-7B model, where LoRA exceeds the performance of other baselines, \ours~further enhances accuracy by 3.7\%, outstripping ChatGPT's accuracy levels. Conversely, for LLaMA-13B, where LoRA's effectiveness is inferior to the Parallel adapter, \ours~achieves superior accuracy over LoRA by 1\% and comparable accuracy to the Parallel adapter, with only a quarter of the trainable parameters required by the Parallel adapter and without adding any extra inference overhead as the Parallel adapter. Additionally, DoRA consistently surpasses LoRA on both LLaMA2-7B and LLaMA3-8B by 2.1\% and 4.4\%, respectively. Furthermore, $\text{\ours}^{\dagger}$ exceeds LoRA's performance on LLaMA-7B by 2.8\%, on LLaMA-13B by 1\%, on LLaMA2-7B by 2.9\%, and on LLaMA3-8B by 4.2\%, despite having only half as many trainable parameters as LoRA. This outcome suggests that the integration of \ours~enhances the learning capability of LoRA, thereby reducing the need for a higher rank to surpass LoRA in terms of accuracy. 

\begin{figure}[h]
\vskip 0.2in
\begin{center}
\centerline{\includegraphics[width=1\columnwidth]{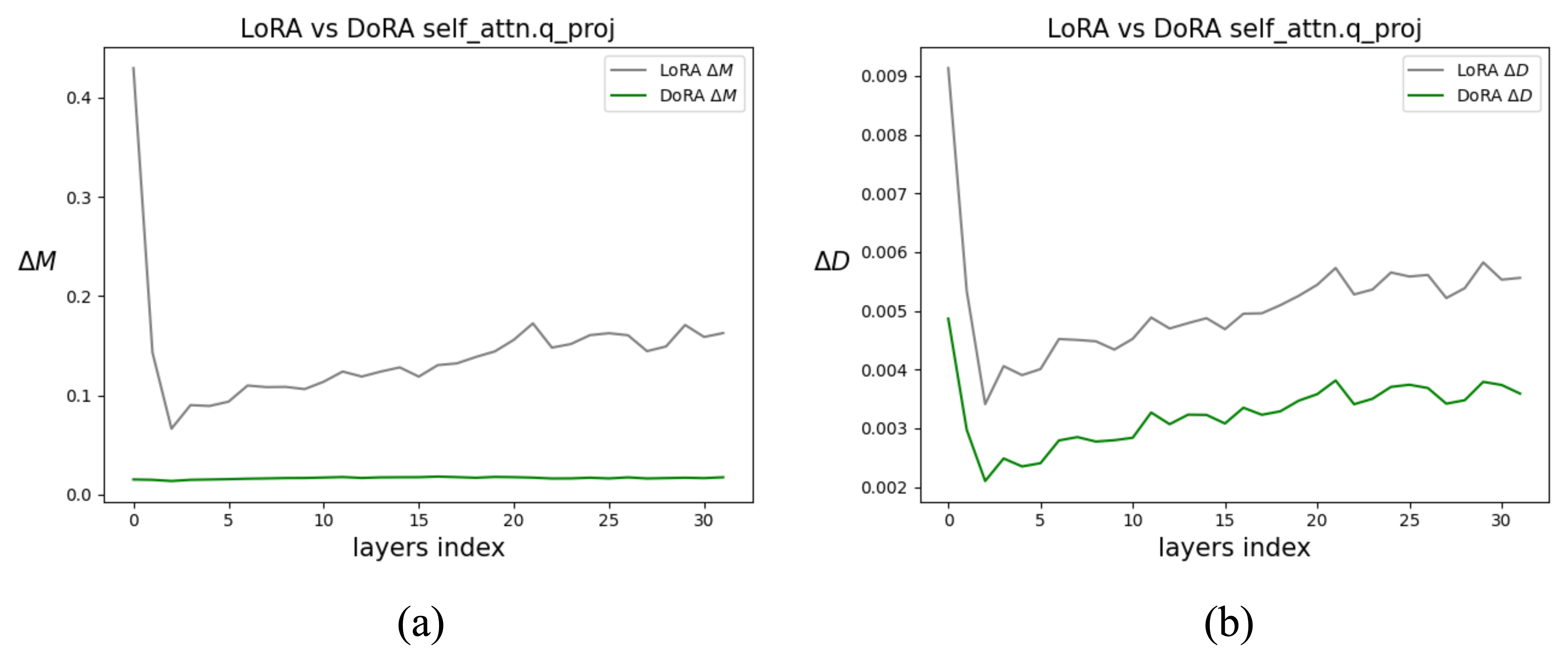}}
\caption{Magnitude (a) and direction (b) difference of LoRA/DoRA and the pre-trained weight of the query matrices across different layers.}
\label{fig:q_proj_delta_m_d}
\end{center}
\vskip -0.5in
\end{figure}

Additionally, in previous sections, we hypothesize that a negative correlation between the magnitude update and directional update is more optimal than a positive correlation. This is because pre-trained weights already contain substantial knowledge suitable for downstream tasks, and a larger magnitude or direction alteration alone is sufficient for downstream adaptation. To further validate our hypothesis, we used LLaMA2-7B fine-tuned with DoRA/LoRA on commonsense reasoning datasets as a case study. We visualized the magnitude ($\Delta M$) and directional difference ($\Delta D$) between the DoRA/LoRA weights and the pre-trained model weights across different modules and layers. In Figure \ref{fig:q_proj_delta_m_d} (a) and (b), we observe that the DoRA fine-tuned weights show less deviation from the pre-trained weights in both magnitude and direction, while the differences for the LoRA fine-tuned weights are significantly larger. Coupled with the experimental results that DoRA significantly outperforms LoRA, we can conclude that our earlier hypothesis is valid: a robust foundation model does not require significant alterations for effective downstream adaptation and having the ability to perform more fine-grained magnitude and directional update explains the superiority of DoRA over LoRA. We leave the visualization of the value and key weight matrices in the appendix.

\subsection{Image/Video-Text Understanding}
\begin{table}[ht]
\centering
\caption{The multi-task evaluation results on VQA, GQA, $\text{NVLR}^2$ and COCO Caption with the VL-BART backbone.}
\vskip 0.1in
\setlength{\tabcolsep}{1.2mm}
\resizebox{0.48\textwidth}{!}{
\begin{tabular}{ccccccc}
\toprule
\textbf{Method} & \# \textbf{Params (\%)} & \textbf{$\text{VQA}^{\text{v2}}$} & \textbf{GQA}& $\textbf{NVLR}^2$& \textbf{COCO Cap} & \textbf{Avg.}\\ \midrule
FT & 100 & 66.9 & 56.7 & 73.7 & 112.0 & 77.3 \\ \hline
LoRA & 5.93 & 65.2 & 53.6 & 71.9 & 115.3 & 76.5 \\ 
\ours~(Ours) & 5.96 & 65.8 & 54.7 & 73.1 & 115.9 & \textbf{77.4} \\ 
\bottomrule
\end{tabular}}
\label{tab:vlbart_image}
\vskip -0.1in
\end{table}

\begin{table}[ht]
\centering
\caption{The multi-task evaluation results on TVQA, How2QA, TVC, and YC2C with the VL-BART backbone.}
\vskip 0.1in
\setlength{\tabcolsep}{1.2mm}
\resizebox{0.48\textwidth}{!}{
\begin{tabular}{ccccccc}
\toprule
\textbf{Method} & \# \textbf{Params (\%)} & \textbf{TVQA} & \textbf{How2QA}& \textbf{TVC}& \textbf{YC2C} & \textbf{Avg.}\\ \midrule
FT & 100 & 76.3 & 73.9 & 45.7 & 154 & 87.5 \\ \hline
LoRA & 5.17 &  75.5 & 72.9 & 44.6 & 140.9 & 83.5 \\ 
\ours~(Ours) & 5.19 & 76.3 & 74.1 & 45.8 & 145.4 & \textbf{85.4} \\ \bottomrule
\end{tabular}}
\label{tab:vlbart_video}
\vskip -0.1in
\end{table}
Having shown that \ours~can consistently achieve better accuracy on fine-tuning LLM, we would like to see if \ours~can remain competitive on multi-modality fine-tuning tasks. We compare \ours~with LoRA and full fine-tuning on VL-BART which comprises a vision encoder (CLIP-ResNet101~\cite{radford2021clip}) and an encoder-decoder language model ($\text{BART}_{\text{Base}}$~\cite{lewis2019bart}) across four different image-text tasks: $\text{VQA}^{\text{v2}}$~\cite{goyal2017vaqv2} and GQA~\cite{hudson2019gqa} for visual question answering, $\text{NLVR}^2$~\cite{suhr2019nlvr2} for visual reasoning, and MSCOCO~\cite{chen2015mscoco} for image captioning, and four different video-text tasks from the VALUE~\cite{li2021value} Benchmark: TVQA \cite{lei2019tvqa} and How2QA \cite{li2020hero} for video question answering, TVC \cite{lei2020tvr} and YC2C \cite{zhou2017automatic} for video captioning. 

We follow the same framework as \cite{sung2022vladapter} and fine-tuned VL-BART within a multi-task framework for both image/video-text tasks. We adopt the same setup as that of LoRA outlined in \cite{sung2022vladapter} when applying \ours. See Table \ref{tab:vlbart_hyperparameters} for the complete hyperparameters. The result of LoRA and FT for both image/video-text tasks is directly quoted from \cite{sung2022vladapter}. We can see that \ours~uniformly surpasses LoRA in accuracy while maintaining a similar count of trainable parameters in both Table \ref{tab:vlbart_image} and Table \ref{tab:vlbart_video}. In particular, \ours~exceeds LoRA's performance by nearly 1\% in image-text understanding tasks, reaching the accuracy level of FT. Moreover, \ours~achieves roughly 2\% higher accuracy than LoRA in video-text understanding tasks.

\subsection{Visual Instruction Tuning}
\begin{table}[h]
\centering
\caption{Visual instruction tuning evaluation results for LLaVA-1.5-7B on a wide range of seven vision-language tasks. We directly use checkpoints from \cite{liu2023llava} to reproduce their results.}
\vskip 0.1in
\small
\begin{tabular}{ccc}
\toprule
\textbf{Method} & \# \textbf{Params}(\%)  & \textbf{Avg.}\\ \midrule
FT & 100 &  66.5 \\ \hline
LoRA & 4.61 & 66.9 \\ 
\ours~(Ours) & 4.63 & \textbf{67.6} \\ 
\bottomrule
\end{tabular}
\label{tab:llava}
\end{table}

We further scale up the model size and compare \ours~to LoRA and FT on the visual instruction tuning tasks with LLaVA-1.5-7B \cite{liu2023llava} which is composed of a language model, Vicuna-1.5-7B~\cite{peng2023vicuna}, and a vision encoder, CLIP ViT-L/336px~\cite{radford2021clip}. The training datasets contain several datasets from VQA \cite{goyal2017vaqv2, hudson2019gqa, marino2019okvqa, schwenk2022aokvqa}, OCR \cite{mishra2019ocrvqa, sidorov2020textcaps}, region-level VQA \cite{kazemzadeh-etal-2014-referitgame, krishna2016visual, mao2016generation}, visual conversation \cite{liu2023llava}, and language conversation data. We follow the setting of \cite{liu2023llava} to filter the training data and construct the tunning prompt format. For a fair comparison, \ours~follows the same configuration as the LoRA configuration provided by \cite{liu2023llava}. The fine-tuned models are then evaluated on seven vision-language benchmarks: $\text{VQA}^{\text{v2}}$ \cite{goyal2017vaqv2}, \text{GQA} \cite{hudson2019gqa}, \text{VisWiz} \cite{gurari2018vizwiz} \text{SQA} \cite{lu2022learn}, $\text{VQA}^{\text{T}}$ \cite{singh2019vqa}, \text{POPE} \cite{li2023evaluating}, and \text{MMBench} \cite{liu2023mmbench}.

From Table \ref{tab:llava}, we can observe that the average accuracy of LoRA already surpasses FT, which could imply that FT might be experiencing issues with overfitting.
Given that \ours~is designed to enhance LoRA's performance to more closely resemble that of FT, in scenarios where FT is inferior to LoRA, \ours's improvement over LoRA might not be as pronounced as observed in other experiments where FT usually outperforms LoRA. Nonetheless, \ours~still demonstrates superior performance over both LoRA and FT, with an average improvement of 0.7\% over LoRA and 1.1\% over FT. See Table \ref{tab:llava_hyperparamters} for the hyperparameters setting and Table \ref{tab:llava_appendix} for the score of each evaluation benchmark.

\subsection{Compatibility of \ours~with other LoRA variants}
\label{sec:elora}
\begin{table}[h]
\centering
\caption{Average scores on MT-Bench assigned by GPT-4 to the answers generated by fine-tuned LLaMA-7B/LLaMA2-7B.}
\vskip 0.1in
\setlength{\tabcolsep}{1.2mm}
\resizebox{0.43\textwidth}{!}{
\begin{tabular}{ccccccc}
\toprule
\textbf{Model} & \textbf{PEFT Method} & \# \textbf{Params (\%)} & \textbf{Score}\\ \midrule
\multirow{4}{*}{LLaMA-7B} & LoRA & 2.31 & 5.1  \\ 
 & \ours~(Ours) & 2.33 & \textbf{5.5}  \\ 
 & VeRA & 0.02 & 4.3  \\ 
 & \ourss~(Ours) & 0.04 & \textbf{5.0}  \\ \midrule
\multirow{4}{*}{LLaMA2-7B} & LoRA & 2.31 & 5.7  \\ 
 & \ours~(Ours) & 2.33 & \textbf{6.0} \\ 
 & VeRA & 0.02 & 5.5  \\ 
 & \ourss~(Ours) & 0.04 & \textbf{6.0} \\ \bottomrule
\end{tabular}}
\label{tab:elora}
\end{table}
Recall from Equation.(\ref{eq:lora}) that $\Delta W$ can be adapted by different LoRA variants. With \ours, the concept of incremental directional update $\Delta V$ introduced in Equation.(\ref{eq:MADoRA}) can likewise be replaced with alternative LoRA variants. In this section, we select VeRA \cite{kopiczko2023vera} as a case study to explore \ours's compatibility with other LoRA variants. VeRA suggests freezing a unique pair of random low-rank matrices to be shared across all layers, employing only minimal layer-specific trainable scaling vectors to capture each layer's incremental updates. This approach allows VeRA to reduce trainable parameters significantly by 10x compared to LoRA, with only a minimal impact on accuracy. 
We apply VeRA for the directional update in \ours~and name such combination \ourss. We assess the effectiveness of both \ourss~and \ours~compared to VeRA and LoRA across LLaMA-7B and LLaMA2-7B, focusing on instruction tuning with the 10K subset of cleaned Alpaca dataset \cite{alpaca}. We utilize the official implementation of VeRA to obtain the results of VeRA and LoRA and fine-tune the model with \ourss~and \ours~using the identical training settings as VeRA and LoRA (see Table~\ref{tab:elora_hyperparamters} in the appendix for more details). The performance of the fine-tuned models is then evaluated on the MT-Bench benchmark \cite{zheng2023judging} by generating model responses to a pre-defined set of 80 multi-turn questions. These responses are then evaluated by GPT-4, which reviews each answer and assigns a numerical score out of 10. 

Table \ref{tab:elora} presents the average scores for \ourss, \ours, VeRA, and LoRA, demonstrating that our proposed method exhibits consistent improvements over VeRA and LoRA for both LLaMA-7B and LLaMA2-7B. 
This effectively showcases the compatibility of \ours~with VeRA. In particular, \ourss~merges the advantageous qualities of \ours~and VeRA, attaining scores that are on par with or even surpass those of LoRA, yet with significantly fewer parameters. For example, \ourss~outperforms VeRA by 0.7/0.5 points and achieves the same level of accuracy as LoRA on LLaMA-7B and \ours~on LLaMA2-7B, respectively. Additionally, we present a selection of questions chosen from MT-Bench, accompanied by the responses from LLaMA2-7B fine-tuned using \ourss~and VeRA in the appendix (Table~\ref{tab:elora_example1} and \ref{tab:elora_example2}) where we can observe that the answers given by \ourss~tend to be more precise and structural.

\begin{figure}[ht]
\vskip 0.2in
\begin{center}
\centerline{\includegraphics[width=\columnwidth]{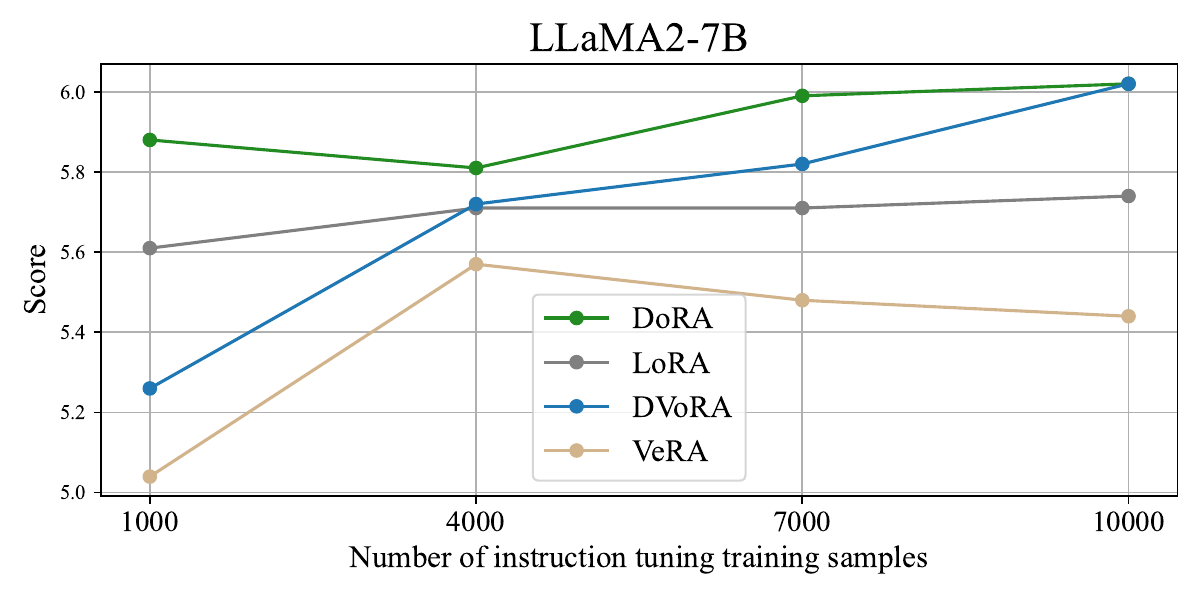}}
\caption{Performance of fine-tuned LLaMA2-7B on MT-Bench using different numbers of Alpaca training samples.}
\label{fig:num_datas}
\end{center}
\vskip -0.2in
\end{figure}

Next, to further assess \ours's ability to remain competitive under varying amounts of training data, considering that in practical situations, access to extensive fine-tuning datasets is frequently limited. We compare \ours~to LoRA and \ourss~to VeRA for fine-tuning LLaMA2-7B/LLaMA-7B with a range of instruction-tuning sample sizes, specifically {1000, 4000, 7000, 10000}, with 10000 being the setting of \cite{kopiczko2023vera}. We visualize the average performance of each method on LLaMA2-7B in Figure~\ref{fig:num_datas}, and on LLaMA-7B in Figure~\ref{fig:num_datas_llama7b} in the appendix. The result shows that \ours~and \ourss~consistently outperform LoRA and VeRA across all training sample sizes. For instance, with 7000 training samples, \ours~and \ourss~surpass LoRA and VeRA by margins of 0.3 and 0.33, respectively. Even when the sample size is reduced to 1000, \ours~and \ourss~maintain their lead with advantages of 0.29 and 0.22 over LoRA and VeRA, respectively. This demonstrates that our methods persistently enhance performance over LoRA and VeRA, regardless of the training sample volume.

\subsection{Robustness of \ours~towards different rank settings}
\begin{figure}[ht]
\vskip 0.2in
\begin{center}
\centerline{\includegraphics[width=\columnwidth]{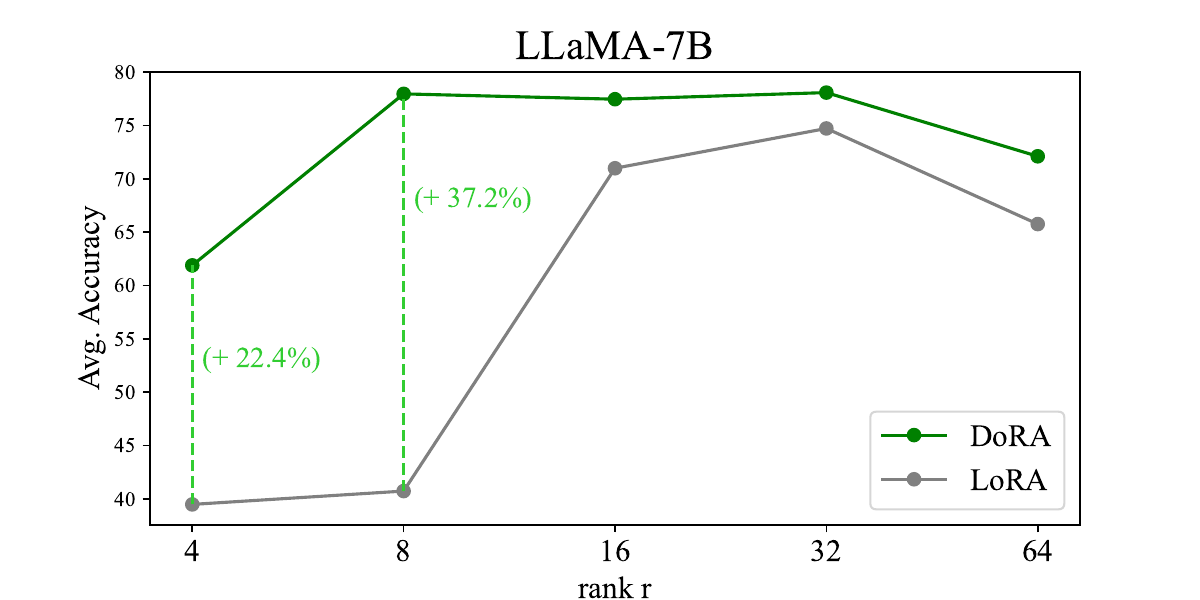}}
\caption{Average accuracy of LoRA and \ours~for varying ranks for LLaMA-7B on the commonsense reasoning tasks.}
\label{fig:dif_rank}
\end{center}
\vskip -0.2in
\end{figure}
This section explores the impact of various rank configurations on \ours~and LoRA by adjusting $r$ within the set \{4, 8, 16, 32, 64\} and assessing the performance of the fine-tuned LLaMA-7B on commonsense reasoning tasks as outlined in Sec~\ref{sec:llama_commonsense}. The average accuracies of LoRA and \ours~across different ranks are depicted in Figure~\ref{fig:dif_rank}, with detailed numbers presented in Table \ref{tab:diff_rank}. From Figure~\ref{fig:dif_rank}, we can observe that \ours~consistently surpasses LoRA across all rank configurations. Notably, the performance gap widens for ranks below 8, where LoRA's average accuracies drop to 40.74\% for $r=8$ and 39.49\% for $r=4$. In contrast, \ours~retains a notable accuracy of 77.96\% for $r=8$ and 61.89\% for $r=4$, demonstrating its resilience and consistently superior performance over LoRA regardless of the rank setting.

\subsection{Tuning Granularity Analysis}
\begin{table}
\caption{Accuracy comparison of LLaMA 7B/13B with two different tuning granularity of \ours. Columns \textbf{m} and \textbf{V} designate the modules with tunable magnitude and directional components, respectively. Each module is represented by its first letter as follows: (Q)uery, (K)ey, (V)alue, (O)utput, (G)ate, (U)p, (D)own.}
\vskip 0.1in
\setlength{\tabcolsep}{0.12mm}
\centering
\tiny
\resizebox{0.46\textwidth}{!}{
\begin{tabular}{cccccc}
\toprule
\textbf{Model} & \textbf{PEFT Method} & \# \textbf{Params (\%)} & \textbf{m}  & \textbf{V} & \textbf{Avg.} \\ \hline
\multirow{3}{*}{LLaMA-7B}    & LoRA & 0.83 & - & - & 74.7  \\ 
&\ours~(Ours) & 0.84 & QKVUD& QKVUD & 78.1  \\ 
& \ours~(Ours) & 0.39 & QKVOGUD & QKV & 77.5  \\  \midrule

\multirow{3}{*}{LLaMA-13B}    & LoRA & 0.67 & - & - & 80.5 \\
& \ours~(Ours) & 0.68 & QKVUD & QKVUD & 81.5  \\ 
& \ours~(Ours) & 0.31 & QKVOGUD & QKV & 81.3  \\ \bottomrule
\end{tabular}}
\label{tab:llama_commonsense_tunning_granularity}
\vskip -0.1in
\end{table}

The visualization in Figure~\ref{fig:q_proj_analysis} indicates that significant changes in magnitude often result in relatively smaller directional changes. Given this observation and the fact that directional updates account for most of the trainable parameters, it prompts an investigation into whether it is possible to decrease the number of trainable parameters by updating only the magnitude components of specific modules while continuing to update both the magnitude and directional components for the remaining linear modules.

Our findings indicate that, in contrast to the original configuration suggested for LoRA in \cite{hu2023llmadapters}, which requires updates to both the Multi-head Attention and MLP layers for optimal performance, \ours~can already achieve superior accuracy by updating only the directional and magnitude components of the multi-head layers and the magnitude of the MLP layers. Specifically, as shown in Table \ref{tab:llama_commonsense_tunning_granularity}, by updating the directional and magnitude components of the QKV modules and only the magnitude of the rest of the layers, \ours~surpasses LoRA by 2.8\% on LLaMA-7B and 0.8\% on LLaMA-13B, while utilizing only less than half of the trainable parameters compared to LoRA.

\section{Broader Impacts}
\subsection{QDoRA: Enhancements to QLoRA}
\begin{figure}[ht]
\vskip 0.2in
\begin{center}
\centerline{\includegraphics[width=1\columnwidth]{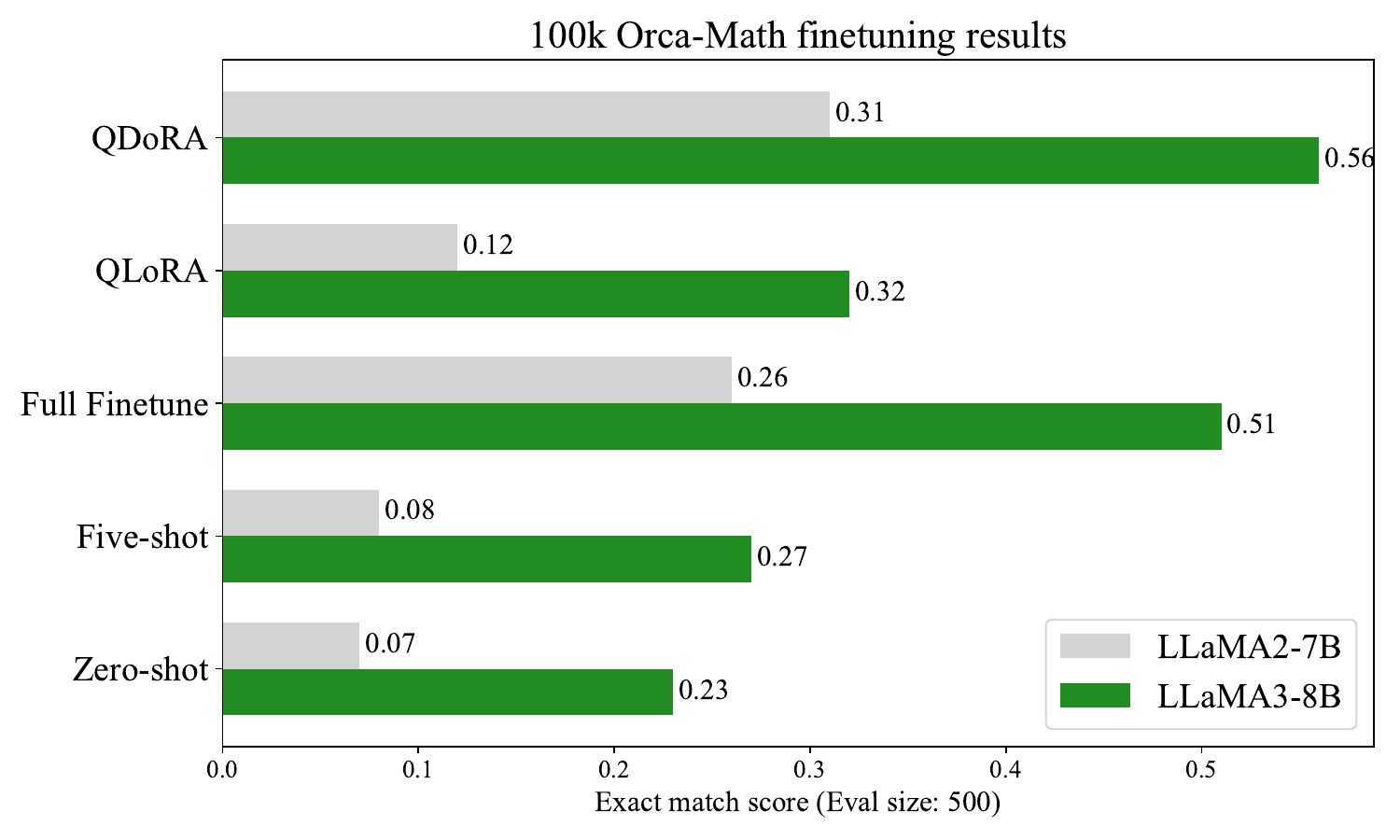}}
\caption{Accuracy comparison of LLaMA2-7B/LLaMA3-8B with QDoRA, QLoRA and FT on Orca-Math~\cite{mitra2024orcamath}.}
\label{fig:qdora}
\end{center}
\vskip -0.2in
\end{figure}
While finetuning LLMs with PEFT significantly reduces training memory overhead, a considerable amount of GPU memory is still required to initially load the model weights onto the GPUs. To further decrease the memory demands of finetuning, QLoRA \cite{NEURIPS2023_qlora} suggests quantizing the pretrained model to 4-bit and finetuning LoRA on top of the frozen low-bit backbone. With our porposed DoRA, which narrows the gap between LoRA and FT, it is natural to also explore whether DoRA can enhance the accuracy of LoRA within the QLoRA framework. Recently, \cite{QDoRA} launch a project that substitutes the LoRA component in QLoRA with DoRA, dubbing it QDoRA, and incorporate the training pipeline with Fully Sharded Data Parallel (FSDP) \cite{zhao2023pytorch} to enable model splitting and parallel training across multiple GPUs. 

They conducted experiments on fine-tuning LLaMA2-7B/LLaMA3-8B using the Orca-Math\cite{mitra2024orcamath} dataset with QDoRA, QLoRA, and FT. The training set included 100k samples, with 500 reserved for evaluation using the exact match score as the metric. In addition to the fine-tuned models, they also reported results from zero-shot, few-shot, and FT with post-training quantization (PTQ), where the FT model is quantized to the BnB NF4 format after training. According to Figure~\ref{fig:qdora}, QDoRA not only significantly surpasses QLoRA by 0.19/0.23 on LLaMA2-7B and LLaMA3-8B, but it also slightly outperforms FT on both models, while using considerably less memory. This indicates that QDoRA can effectively combines the parameter efficiency of QLoRA with the more granular optimization of full finetuning. These initial findings suggest that QDoRA holds considerable promise and could hugely benefit the opensoure community by substantially lowering the GPU memory requirements for fine-tuning large language models.

\subsection{Text-to-Image Generation}
Recently, as diffusion models have expanded in size, LoRA has become a popular method for efficiently fine-tuning large stable diffusion models. In this section, we aim to explore whether DoRA's advantages over LoRA extend to the task of text-to-image generation. We follow the training pipeline of DreamBooth~\cite{ruiz2023dreambooth} for fine-tuning SDXL~\cite{podell2023sdxl}, utilizing the advanced training scripts developed by HuggingFace. The hyperparameter settings for LoRA and DoRA are kept the same, and we fine-tune the model using two challenging datasets: 3D icons and Lego sets. The sample seeds for generating the images are kept the same for LoRA and DoRA for fair comparison. The generated images are shown in Figure \ref{fig:sdxl_3d} and \ref{fig:sdxl_lego} in the appendix. The results indicate that DoRA achieves significantly better personalization than LoRA when using the same training settings, and more accurately reflects the training targets. For example, in Figure \ref{fig:sdxl_3d}, the first sub-figure of DoRA's output features a unique round square around the image, which is a feature common to all the training targets. In contrast, this feature is absent in all the LoRA outputs. A similar observation could be found with the Lego training targets, where only the DoRA outputs consistently incorporate the Lego logo in the generated images.

\section{Conclusion}
In this work, we first conduct a novel weight decomposition analysis to reveal the distinct learning patterns between LoRA and FT. Building on these insights, we introduce \ours, a fine-tuning method that is compatible with LoRA and its variants and exhibits a closer resemblance to FT's learning behavior. \ours~consistently outperforms LoRA across various fine-tuning tasks and model architectures. Specifically, \ours~improves upon LoRA in commonsense reasoning and visual instruction tuning tasks. Furthermore, \ours~also shows compatibility with VeRA on the Alpaca instruction tuning task. Moreover, \ours~can be considered as a costless alternative to LoRA, as its decomposed magnitude and direction components can be merged back into the pre-trained weight after the training, ensuring that there is no extra inference overhead. For future work, we wish to explore the generalizability of \ours~in domains beyond language and vision, particularly in the field of audio.

\newpage
\section*{Acknowledgements}
We extend our gratitude to Benjamin Bossan, Younes Belkada, and Sourab Mangrulkar from Hugging Face for their assistance in integrating DoRA into the PEFT package, thus making our work more accessible to the broader public. We thank Kerem Turgutlu, Jonathan Whitaker, and Jeremy Howard from Answer.AI for their work on the implementation and experiments of QDoRA/FSDP, which makes fine-tuning of large language models with DoRA on consumer GPUs a lot more feasible. We also thank Sebastian Raschka for his well-written tutorial on DoRA which offers a thorough overview of the background knowledge necessary to comprehend DoRA.

\section*{Impact Statement}
This paper presents work whose goal is to advance the field of Machine Learning. There are many potential societal consequences of our work, none of which we feel must be specifically highlighted here.

\bibliography{example_paper}
\bibliographystyle{icml2024}

\newpage
\appendix
\onecolumn
\section{Appendix}
\subsection{Weight decomposition analysis on the value weight matrix}
\begin{figure}[ht]
\vskip 0.2in
\begin{center}
\centerline{\includegraphics[width=\columnwidth]{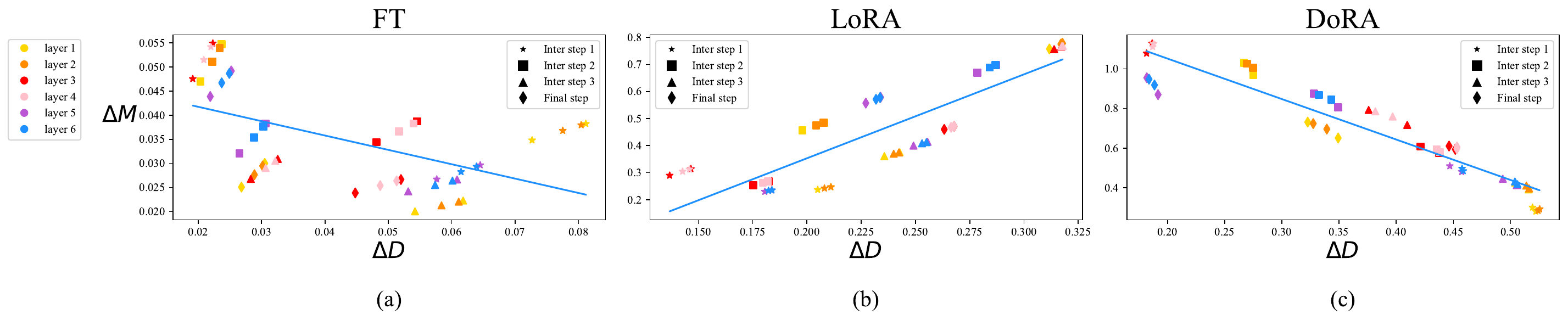}}
\caption{Magnitude and Directional changes of FT (a), LoRA (b), and \ours~(c) of the V weight matrices across different layers and intermediate steps.}
\label{fig:v_proj_analysis}
\end{center}
\vskip -0.2in
\end{figure}

In this section, we illustrate the changes in magnitude and direction within the value weight matrix for FT, LoRA, and \ours~across different training steps and layers, as shown in Figure \ref{fig:v_proj_analysis}. This reveals patterns similar to those seen in the query weight matrix depicted in Figure \ref{fig:q_proj_analysis}, indicating that \ours~is capable of displaying learning behaviors that closely mirror those of FT across various modules.

\subsection{Ablation study for the modification to reduce \ours~training cost}
Table \ref{tab:training_cost} presents the GPU cost and the average accuracy of \ours~with and without the proposed modification for commonsense reasoning tasks and image-text understanding tasks. The results indicate that the modification leads to a training memory reduction of approximately 24.4\% in fine-tuning LLaMA and 12.4\% in VL-BART. Furthermore, the accuracy of \ours~with the modification remains unchanged for VL-BART and shows a negligible difference of only 0.2 compared to \ours~without the modification on LLaMA.

\begin{table}[h]
\caption{GPU cost and accuracy of \ours~with or without the modification on the commonsense reasoning tasks and image-text understanding tasks.}
\vskip 0.1in
\centering
\resizebox{\textwidth}{!}{
\begin{tabular}{ccccccc}
\toprule
\textbf{Model} & \textbf{PEFT Method} & \textbf{ Accumulation steps } & \textbf{Batch Size} & \textbf{GPU Memory Cost (GB)} & \textbf{\# Params (\%)}  & \textbf{Avg.} \\ \hline
\multirow{2}{*}{LLaMA-7B}  & \ours~w/o modification &  \multirow{2}{*}{4} & \multirow{2}{*}{16} &  37.3 & 0.84 & 78.3 \\ 
& \ours & & & 28.2 {\color{teal}(-24.4\%)} & 0.84 & 78.1  \\  \midrule
\multirow{2}{*}{VL-BART}  & \ours~w/o modification &  \multirow{2}{*}{-} & \multirow{2}{*}{300} & 23.4 & 5.96 & 77.3 \\ 
& \ours & & & 20.5 {\color{teal}(-12.4\%)} & 5.96 & 77.4  \\  \midrule


\end{tabular}}
\label{tab:training_cost}
\vskip -0.1in
\end{table}

\newpage
\subsection{Hyperparameters}
\begin{table}[h]
\centering
\caption{Hyperparameter configurations of \ours~for LLaMA-7B/13B, LLaMA2-7B, and LLaMA3-8B on the commonsense reasoning tasks.}
\vskip 0.1in
\small
\begin{tabular}{ccccccccc}
\toprule
\textbf{Hyperparameters (\ours)} & \multicolumn{2}{c}{LLaMA-7B} & \multicolumn{2}{c}{LLaMA-13B} & \multicolumn{2}{c}{LLaMA2-7B} & \multicolumn{2}{c}{LLaMA3-8B} \\ \midrule
Rank r & 16 & 32 & 16 & 32 & 16 & 32 & 16 & 32 \\ 
$\alpha$ & 32 & 64 & 32 & 64 & 32 & 64 & 32 & 64 \\ 
Dropout & \multicolumn{8}{c}{0.05} \\ 
Optimizer & \multicolumn{8}{c}{AdamW}  \\ 
LR & 2e-4 & 1e-4 & 3e-4 & 2e-4 & 2e-4 & 2e-4 & 1e-4 & 1e-4  \\
LR Scheduler & \multicolumn{8}{c}{Linear} \\
Batch size & \multicolumn{8}{c}{16} \\
Warmup Steps & \multicolumn{8}{c}{100} \\
Epochs & \multicolumn{8}{c}{3} \\
Where & \multicolumn{8}{c}{Q,K,V,Up,Down} \\
\bottomrule
\end{tabular}
\label{tab:llama_commonsense_hyperparameters}
\vskip -0.1in
\end{table}

\begin{table}[h]
\centering
\caption{Hyperparameter configurations of \ours~for fine-tuning VL-Bart on image/video-text tasks.}
\vskip 0.1in
\small
\begin{tabular}{ccc}
\toprule
\textbf{Hyperparameters (\ours)} & image-text & video-text \\ \midrule
Rank r & \multicolumn{2}{c}{128}  \\ 
$\alpha$ & \multicolumn{2}{c}{128}  \\ 
Dropout & \multicolumn{2}{c}{0.0} \\ 
Optimizer & \multicolumn{2}{c}{AdamW}  \\ 
LR & 1e-3 & 3e-4  \\
LR Scheduler & \multicolumn{2}{c}{Linear} \\
Batch size & 300 & 40 \\
Warmup ratio & \multicolumn{2}{c}{0.1} \\
Epochs & 20 & 7 \\
Where & \multicolumn{2}{c}{Q,K} \\
\bottomrule
\end{tabular}
\label{tab:vlbart_hyperparameters}
\vskip -0.1in
\end{table}

\begin{table}[h]
\centering
\caption{Hyperparameter configurations of \ours~and LoRA for fine-tuning LLaVA-1.5-7B with visual instruction tuning datasets.}
\vskip 0.1in
\small
\begin{tabular}{ccc}
\toprule
\textbf{Hyperparameters} & \ours & LoRA \\ \midrule
Rank r & \multicolumn{2}{c}{128} \\ 
$\alpha$ & \multicolumn{2}{c}{256} \\ 
Dropout & \multicolumn{2}{c}{0.05} \\ 
Optimizer & \multicolumn{2}{c}{AdamW}  \\ 
LR & \multicolumn{2}{c}{2e-4} \\
LR Scheduler & \multicolumn{2}{c}{Cosine decay} \\
Batch size & \multicolumn{2}{c}{16} \\
Warmup ratio & \multicolumn{2}{c}{0.03} \\
Epochs & \multicolumn{2}{c}{1} \\
Where & \multicolumn{2}{c}{Q,K,V,O,Up,Down,Gate} \\
\bottomrule
\end{tabular}
\label{tab:llava_hyperparamters}
\vskip -0.1in
\end{table}

\begin{table}[ht]
\centering
\caption{Hyperparameter configurations of \ours~and \ourss~for fine-tuning LLaMA-7B and LLaMA2-7B with cleaned Alpaca dataset.}
\vskip 0.1in
\small
\begin{tabular}{ccc}
\toprule
\textbf{Hyperparameters (\ours)} & \multicolumn{1}{c}{LLaMA-7B} & \multicolumn{1}{c}{LLaMA2-7B} \\ \midrule
Rank r &  \multicolumn{2}{c}{64} \\ 
Dropout & \multicolumn{2}{c}{0.0} \\ 
Optimizer & \multicolumn{2}{c}{AdamW}  \\ 
LR & \multicolumn{2}{c}{4e-4}   \\
LR Scheduler & \multicolumn{2}{c}{Cosine} \\
Batch size & \multicolumn{2}{c}{4} \\
Accumulation Steps & \multicolumn{2}{c}{4} \\
Warmup ratio & \multicolumn{2}{c}{0.1} \\
Epochs & \multicolumn{2}{c}{1} \\
Where & \multicolumn{2}{c}{Q,K,V,O,Up,Down,Gate} \\
\bottomrule
\end{tabular}
\begin{tabular}{ccc}
\toprule
\textbf{Hyperparameters (\ourss)} & \multicolumn{1}{c}{LLaMA-7B} & \multicolumn{1}{c}{LLaMA2-7B} \\ \midrule
Rank r &  \multicolumn{2}{c}{1024} \\ 
Dropout & \multicolumn{2}{c}{0.0} \\ 
Optimizer & \multicolumn{2}{c}{AdamW}  \\ 
LR & \multicolumn{2}{c}{4e-3}   \\
LR Scheduler & \multicolumn{2}{c}{Cosine} \\
Batch size & \multicolumn{2}{c}{4} \\
Accumulation Steps & \multicolumn{2}{c}{4} \\
Warmup ratio & \multicolumn{2}{c}{0.1} \\
Epochs & \multicolumn{2}{c}{1} \\
Where & \multicolumn{2}{c}{Q,K,V,O,Up,Down,Gate} \\
\bottomrule
\end{tabular}
\label{tab:elora_hyperparamters}
\vskip -0.1in
\end{table}

\clearpage
\newpage
\subsection{Magnitude and Direction difference between DoRA/LoRA fine-tuned weight and the pre-triained weight of LLaMA2-7B for the commonsesne reasoning tasks}
\begin{figure}[ht]
\vskip 0.2in
\begin{center}
\centerline{\includegraphics[width=0.65\columnwidth]{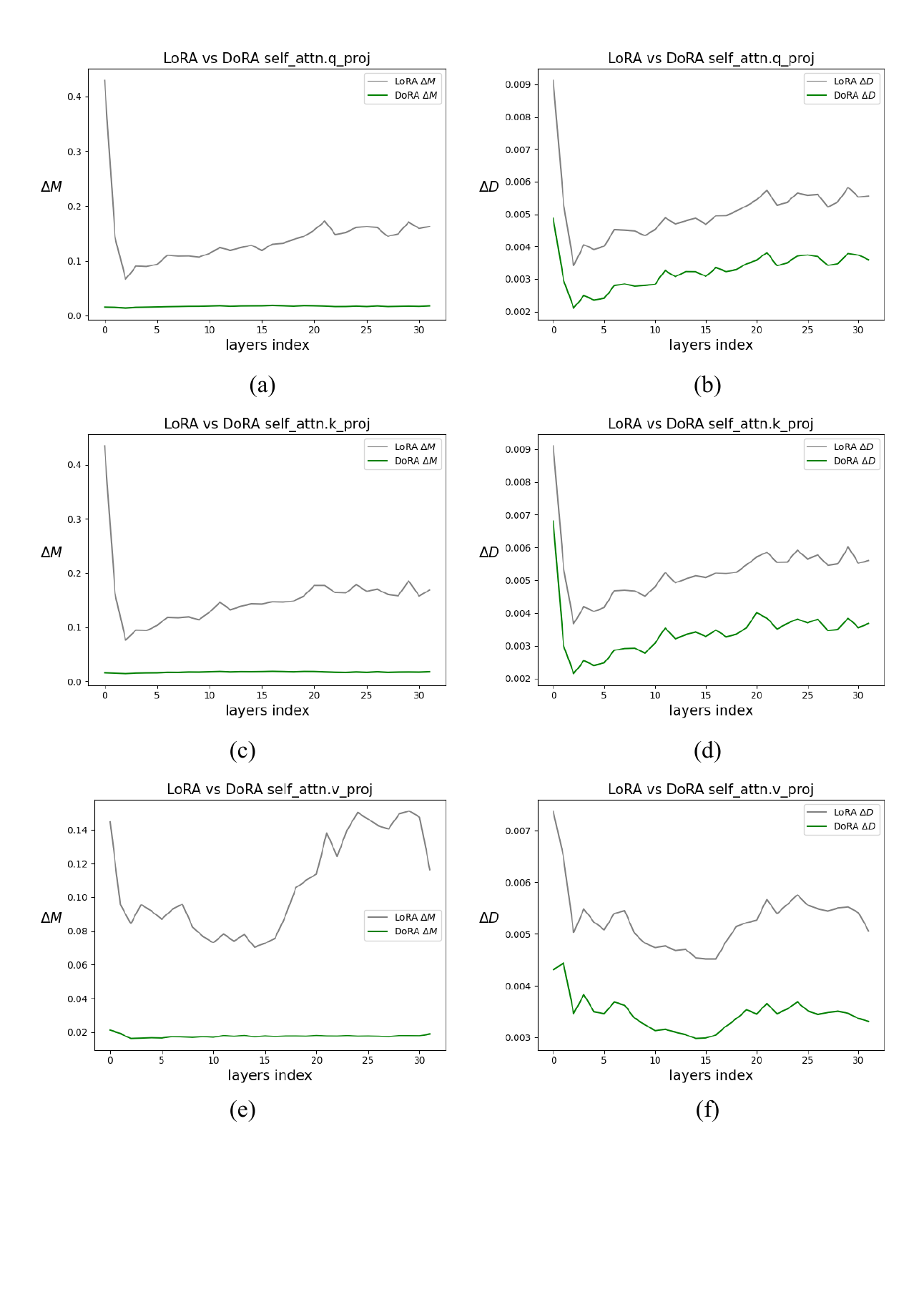}}
\caption{Magnitude and direction differences in the weights of the query, key, and value matrices between LoRA/DoRA fine-tuned models and the pre-trained model across various layers of LLaMA2-7B for the commonsense reasoning tasks.}
\label{fig:all_delta_d_m}
\end{center}
\vskip -0.2in
\end{figure}

Figure \ref{fig:all_delta_d_m} depicts the magnitude and direction differences in the weights of the query, key, and value matrices between LoRA/DoRA fine-tuned models and the pre-trained model across various layers of LLaMA2-7B for the commonsense reasoning tasks. The figure shows that the DoRA fine-tuned weights deviate less from the pre-trained weights in both magnitude and direction, supporting our hypothesis that a robust foundation model does not need substantial changes for effective downstream adaptation.

\clearpage
\newpage
\subsection{Visual instruction tuning evaluation result}
Table \ref{tab:llava_appendix} presents the evaluation result of LLaVA-1.5-7B fine-tuned using \ours, LoRA, and FT with visual instruction tuning data. The evaluation is conducted on an array of seven different vision-language benchmarks. From the table, we can see that \ours~consistently outmatch FT, and can achieve an average accuracy better than LoRA.
\begin{table}[ht]
\centering
\caption{Visual instruction tuning evaluation result of \ours, LoRA, and FT for LLaVA-1.5-7B on a wide range of 7 vision-language tasks.}
\setlength{\tabcolsep}{1.2mm}
\resizebox{0.8\textwidth}{!}{
\begin{tabular}{cccccccccc}
\toprule
\textbf{Method} & \# \textbf{Params} (\%) & $\textbf{VQA}^{\text{v2}}$ & \textbf{GQA}& \textbf{VisWiz}& \textbf{SQA} & $\textbf{VQA}^{\text{T}}$ & \textbf{POPE} & \textbf{MMBench}  & \textbf{Avg.}\\ \midrule
FT & 100 & 78.5 & 61.9 & 50.0 & 66.8 & 58.2 & 85.9 & 64.3 & 66.5 \\ \hline
LoRA & 4.61 & 79.1 & 62.9 & 47.8 & 68.4 & 58.2 & 86.4 & 66.1 & 66.9 \\ 
\ours~(Ours) & 4.63 & 78.6 & 62.9 & 52.2 & 69.9 & 57 & 87.2 & 66.1 & \textbf{67.6} \\ 
\bottomrule
\end{tabular}}
\label{tab:llava_appendix}
\end{table}

\newpage
\subsection{Instruction-tuning evaluation examples}
We sample two questions from the MT-Bench question set and present the respective responses from \ourss~and VeRA, including the scores and comments provided by GPT-4. From these examples, we can observe that responses from \ourss~are generally more accurate and well-organized compared to those from VeRA. For instance, in response to Question 24, \ourss~leverages a well-known superhero story that is familiar to children, whereas VeRA's response begins with an uncertain tone, potentially making it less engaging than \ourss's reply.

\begin{table}[h]
\caption{Question 24 from MT-Bench alongside with response from LLaMA2-7B fine-tuned with \ourss~and VeRA, and the evaluation provided by GPT-4.}
\vskip 0.1in
\begin{tabularx}{\textwidth}{lX}
\hline
Question 24  &  As a superhero, how would you explain your origin story to a curious child?  \\ \midrule
VeRA &   Well, I'm not really sure where I come from. All I know is that one day, I woke up with special powers and abilities that allowed me to help people. I decided to use my powers to fight crime and protect the innocent, and that's how I became a superhero  \\ \midrule
\ourss &  As a superhero, I have the ability to use my super powers to protect people and fight against evil. My origin story begins when I was just a regular kid like you, but one day, I was bitten by a radioactive spider, and it gave me incredible powers. I could climb walls, shoot webs, and see in the dark. I decided to use my powers to help people, and that's how I became your friendly neighborhood Spider-Man.   \\ \midrule
GPT-4 Evaluation & Score: 5 7. Assistant 1's response starts with an uncertain tone, which might not be very engaging for a child. The explanation is quite vague and doesn't offer a clear or imaginative origin story that could captivate a child's interest. It does, however, maintain the theme of using powers for good, which is a positive message. Assistant 2's response is more detailed and uses a familiar superhero origin story (Spider-Man's) that many children can relate to and understand. It's engaging and provides a clear narrative of how an ordinary person can become a superhero. However, it directly references Spider-Man, which is not an original origin story and could be misleading if the assistant is supposed to be a unique superhero. Despite this, the response is more likely to resonate with a child due to its clarity and reliability. Both responses could be improved by providing an original and imaginative origin story that doesn't rely on existing superhero narratives. They should also ensure that the story is simple enough for a child to understand while being engaging and inspiring. \\ \bottomrule
\end{tabularx}
\label{tab:elora_example1}
\end{table}
\begin{table}[hbt!]
\caption{Question 33 from MT-Bench alongside with response from LLaMA2-7B fine-tuned with \ourss~and VeRA, and the evaluation provided by GPT-4.}
\begin{tabularx}{\textwidth}{lX}
\hline
Question 33  &  Why might someone choose to use a paper map or ask for directions instead of relying on a GPS device or smartphone app?  \\ \midrule
VeRA &   There are several reasons why someone might choose to use a paper map or ask for directions instead of relying on a GPS device or smartphone app. Some people may prefer the physical act of navigating with a paper map, as it can be more tactile and interactive. Others may prefer to have a hard copy of the map in case the GPS or app fails or the battery dies. Additionally, some people may not have access to a GPS device or smartphone, or may not want to rely on technology for navigation. Finally, some people may simply prefer the security and peace of mind that comes with knowing exactly where they are and how to get to their destination  \\ \midrule
\ourss &  There are several reasons why someone might choose to use a paper map or ask for directions instead of relying on a GPS device or smartphone app. Here are some of them: 1. Navigation accuracy: GPS devices and smartphone apps rely on satellite signals to pinpoint your location, but this technology is not foolproof. GPS signals can be blocked by tall buildings, tunnels, or other obstacles, resulting in inaccurate or missing information. Paper maps, on the other hand, are based on accurate, up-to-date information that is not subject to these limitations. 2. Cost: GPS devices and smartphone apps can be expensive to purchase and maintain. Paper maps are free to use and do not require an ongoing subscription or data plan. 3. Personal preference: Some people prefer the tactile experience of using a paper map, which can be easier to read and navigate than a screen. Others enjoy the challenge of navigating without technology, which can help them learn more about the area they are visiting. 4. Battery life: GPS devices and smartphone apps can drain batteries quickly, especially when used for navigation. Paper maps do not require power, so they can be used for extended periods without recharging. 5. Offline use: GPS devices and smartphone apps rely on an internet connection to function properly. If there is no signal, the device will not be able to provide accurate information. Paper maps, on the other hand, can be used offline, making them ideal for use in remote or unpopulated areas. In summary, there are several reasons why someone might choose to use a paper map or ask for directions instead of relying on a GPS device or smartphone app. These include navigation accuracy, cost, personal preference, battery life, and offline use.\\ \midrule
GPT-4 Evaluation & Score: 6 8. Assistant 1's response is somewhat brief and lacks specific details that could be helpful to the user. It mentions the tactile nature of paper maps, the possibility of technology failure, lack of access to technology, and a general preference for security and peace of mind. However, it does not delve into the practical considerations such as GPS signal limitations, cost, battery life, or offline use, which are important factors in deciding between paper maps and digital navigation tools. Assistant 2's response is more comprehensive and provides a structured list of reasons, including navigation accuracy, cost, personal preference, battery life, and offline use. Each point is explained clearly, giving the user a better understanding of the potential advantages of paper maps over GPS devices or smartphone apps. The response is relevant, accurate, and detailed, making it more helpful to the user. Both assistants provided relevant and accurate information, but Assistant 2's answer is more detailed and structured, which is why it receives a higher score. The scores reflect the level of detail and the usefulness of the information provided to the user, with Assistant 2 offering a more complete answer to the question  \\ \bottomrule
\end{tabularx}
\label{tab:elora_example2}
\vskip -0.1in
\end{table}

\clearpage
\newpage
\subsection{Robustness towards different numbers of instruction tunning examples}
\begin{figure}[ht]
\vskip 0.2in
\begin{center}
\centerline{\includegraphics[width=0.5\columnwidth]{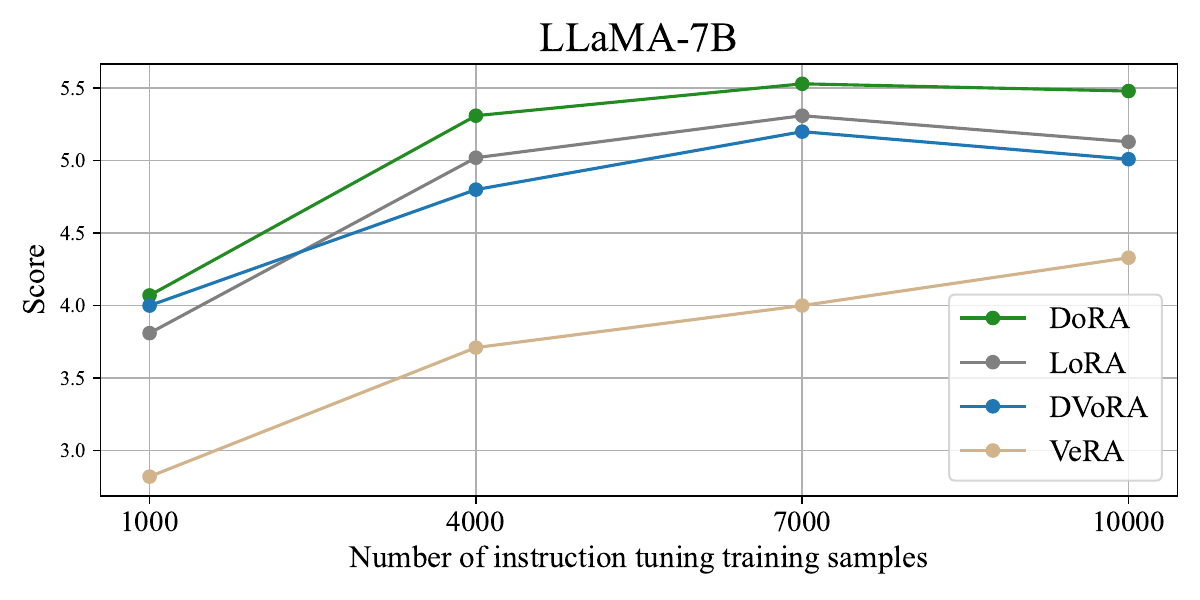}}
\caption{Performance of LLaMA-7B fine-tuned with LoRA, \ours, VeRA, and \ourss~on MT-Bench using different numbers of training samples from Alpaca dataset~\cite{alpaca}.}
\label{fig:num_datas_llama7b}
\end{center}
\vskip -0.2in
\end{figure}

Figure \ref{fig:num_datas_llama7b} depicts the score of \ours, LoRA, \ourss, and VeRA for fine-tuning LLaMA-7B with varying instruction-tuning sample sizes, including 1000, 4000, 7000, and 10000. From the figure, we can observe that similar to that of the result on LLaMA2-7B, both \ours~and \ourss~consistently outperform LoRA and VeRA across the entire range of training sample sizes. 

\newpage
\subsection{Robustness of \ours~towards different rank}
Table \ref{tab:diff_rank} shows a comparison of the average accuracy between LoRA and \ours~method across various rank settings for commonsense reasoning tasks. \ours~consistently outperforms LoRA at all rank settings, with the performance gap widening as the rank decreases. This suggests that our method effectively enhances the learning capacity of LoRA, enabling it to achieve better accuracy with fewer trainable parameters.
\begin{table*}[h]
\caption{Accuracy comparison of LoRA and \ours~with varying ranks for LLaMA-7B on the commonsense reasoning tasks.}
\vskip 0.1in
\setlength{\tabcolsep}{1.2mm}
\centering
\resizebox{0.99\textwidth}{!}{
\begin{tabular}{ccccccccccccc}
\toprule
\textbf{PEFT Method} & \textbf{rank r} & \# \textbf{Params (\%) } & \textbf{BoolQ} & \textbf{PIQA}&\textbf{SIQA}& \textbf{HellaSwag} & \textbf{WinoGrande}& \textbf{ARC-e} & \textbf{ARC-c} & \textbf{OBQA} & \textbf{Avg.} \\ \hline

\multirow{5}{*}{LoRA}    & 4 & 0.10 & 2.3 & 46.1 & 18.3 & 19.7 & 55.2 & 65.4 & 51.9 & 57 & 39.5 \\ 
    & 8 & 0.21 & 31.3 & 57.0 & 44.0 & 11.8 & 43.3 & 45.7 & 39.2 & 53.8 & 40.7 \\
 & 16 & 0.42 & 69.9 & 77.8 & 75.1 & 72.1 & 55.8 & 77.1 & 62.2 & 78.0 & 70.9 \\ 
    & 32 & 0.83 & 68.9 & 80.7 & 77.4 & 78.1 & 78.8 & 77.8 & 61.3 & 74.8 & 74.7 \\ 
    & 64  & 1.64 & 66.7 & 79.1 & 75.7 & 17.6 & 78.8 & 73.3 & 59.6 & 75.2 & 65.8 \\ \hline
\multirow{5}{*}{\ours~(Ours)}    & 4 & 0.11 & 51.3 & 42.2 & 77.8 & 25.4 & 78.8 & 78.7 & 62.5 & 78.6 & 61.9 \\
    & 8 & 0.22 & 69.9 & 81.8 & 79.7 & 85.2 & 80.1 & 81.5 & 65.7 & 79.8 & 77.9 \\ 
 & 16 & 0.43 & 70.0 & 82.6 & 79.7 & 83.2 & 80.6 & 80.6 & 65.4 & 77.6 & 77.5 \\ 
 & 32 & 0.84 & 69.7 &	83.4 &	78.6 &	87.2 &	81.0 &	81.9 &	66.2 &	79.2 &	78.4 \\ 
 & 64 & 1.65 & 69.9 & 81.4 & 79.1 & 40.7 & 80.0 & 80.9 & 65.5 & 79.4 & 72.1 \\ \hline
\end{tabular}}
\label{tab:diff_rank}
\vskip -0.1in
\end{table*}

\clearpage
\newpage
\subsection{Text-to-Image Generation}
\begin{figure}[ht]
\vskip 0.2in
\begin{center}
\centerline{\includegraphics[width=0.6\columnwidth]{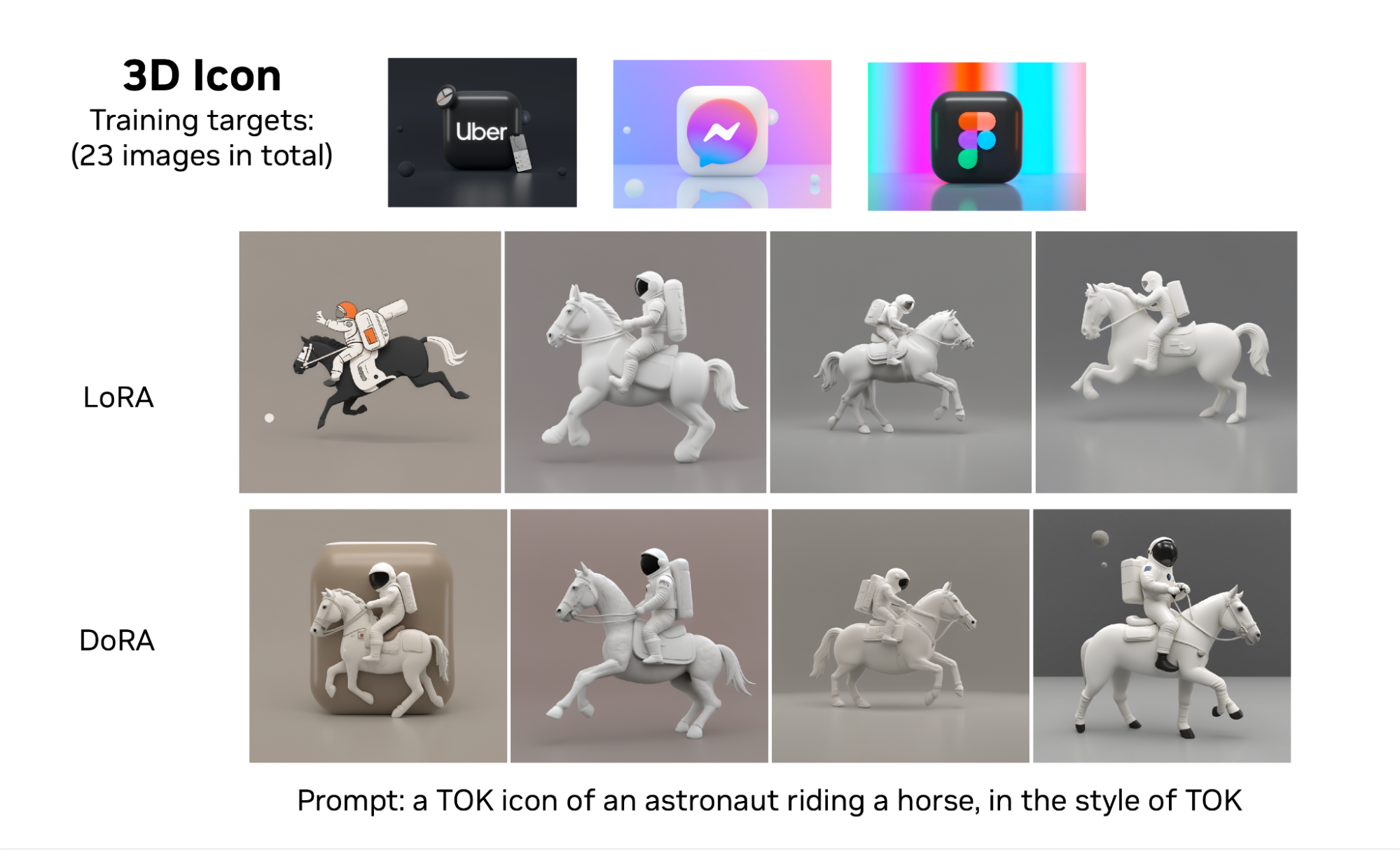}}
\caption{Images generated with SDXL finetuned with LoRA and DoRA on the 3D Icon training sets.}
\label{fig:sdxl_3d}
\end{center}
\vskip -0.2in
\end{figure}

\begin{figure}[ht]
\vskip 0.2in
\begin{center}
\centerline{\includegraphics[width=0.6\columnwidth]{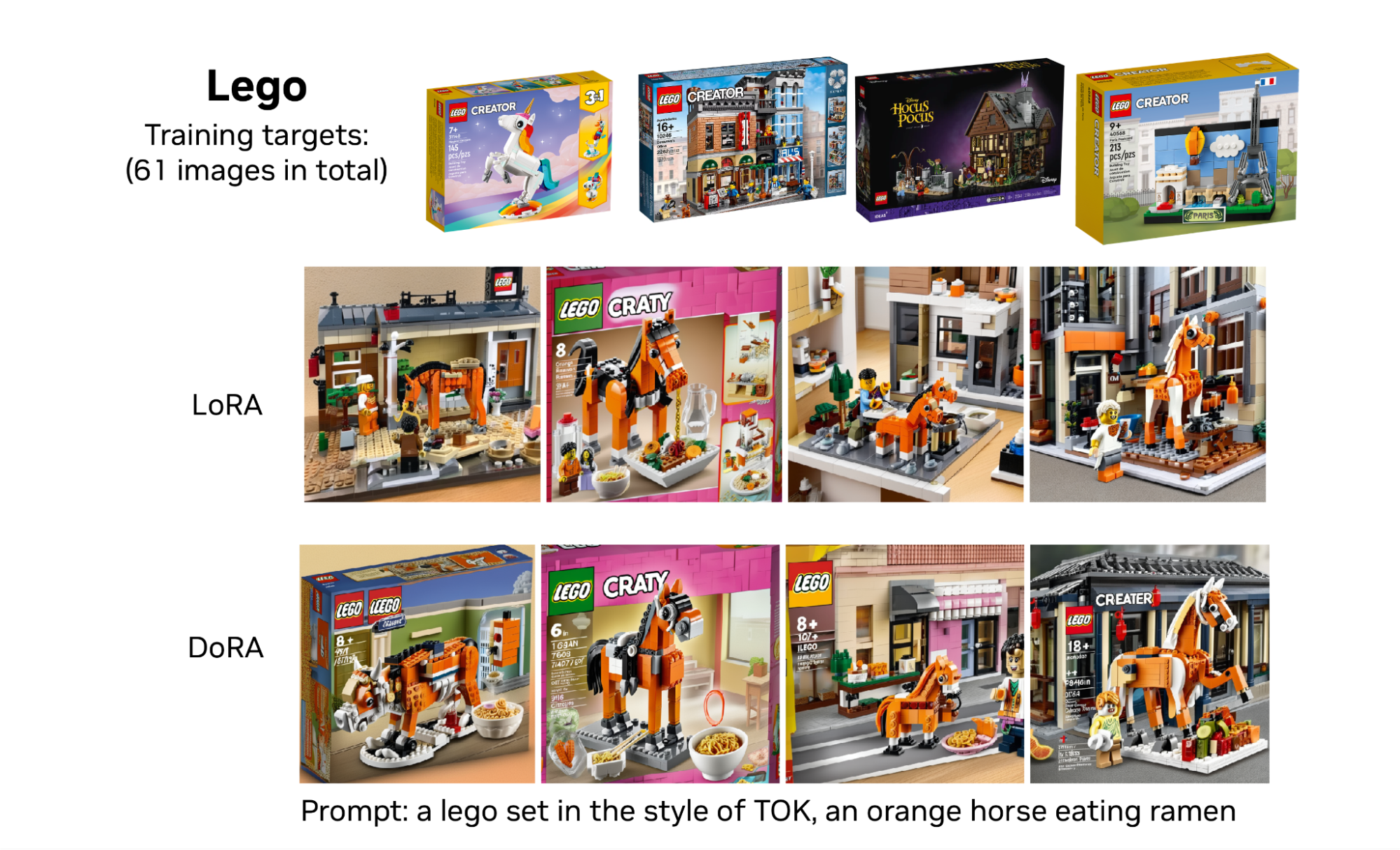}}
\caption{Images generated with SDXL finetuned with LoRA and DoRA on the Lego training sets.}
\label{fig:sdxl_lego}
\end{center}
\vskip -0.2in
\end{figure}

Figures \ref{fig:sdxl_3d} and \ref{fig:sdxl_lego} show the images produced by SDXL fine-tuned with ~\ours  and LoRA via DreamBooth~\cite{ruiz2023dreambooth} personalization techniques on two distinct training sets: 3D Icon\footnote{\url{https://huggingface.co/datasets/linoyts/3d_icon}} and Lego\footnote{\url{https://huggingface.co/datasets/merve/lego_sets_latest}}. The results reveal that DoRA can achieve considerably better personalization than LoRA with identical training configurations, more closely matching the training target.


\end{document}